%% file: Main.tex
\documentclass[preprint,11pt]{elsarticle}
\usepackage{amssymb}
\usepackage{amsmath}
\usepackage{booktabs} 
\usepackage{array}  
\usepackage{pdflscape}
\usepackage{longtable} 
\usepackage{adjustbox} 
\usepackage{lscape} 
\usepackage{geometry} 
\usepackage{graphicx}
\usepackage{tikz}
\usepackage{appendix}
\usepackage{pgf-pie}
\usetikzlibrary{arrows,positioning,shapes.geometric, arrows.meta}
\usepackage{caption}
\usepackage{url}
\usepackage{subcaption}
\usepackage{verbatim}
\usepackage{pgfplots}
\pgfplotsset{compat=1.18}
\usepackage{oldgerm}
\usepackage{lineno}
\usepackage{colortbl}
\usepackage{multirow}
\usepackage{orcidlink}
\usepackage{courier} 
\usepackage{booktabs} 
\usepackage{hyperref}
\usepackage[normalem]{ulem}
\usepackage{comment}

\newcommand{\cyr}[1]{{\color{black} #1}}

\definecolor{inputcolor}{RGB}{77,175,74}    
\definecolor{hiddencolor}{RGB}{55,126,184}  
\definecolor{outputcolor}{RGB}{228,26,28}   
\definecolor{evaluationcolor}{RGB}{152,78,163} 
\definecolor{bordeaux}{RGB}{128, 0, 32}
\definecolor{violet}{RGB}{75, 0, 130}

\definecolor{matlab1}{rgb}{0.000, 0.447, 0.741} 
\definecolor{matlab2}{rgb}{0.850, 0.325, 0.098} 
\definecolor{matlab3}{rgb}{0.929, 0.694, 0.125} 
\definecolor{matlab4}{rgb}{0.494, 0.184, 0.556} 
\definecolor{matlab5}{rgb}{0.466, 0.674, 0.188} 
\definecolor{matlab6}{rgb}{0.301, 0.745, 0.933} 
\definecolor{matlab7}{rgb}{0.635, 0.078, 0.184} 

\definecolor{colTotal}{rgb}     {    0.2422,    0.1504,    0.6603}
\definecolor{colThermal}{rgb}   {    0.2780,    0.3556,    0.9777}
\definecolor{colHydraulic}{rgb} {    0.1540,    0.5902,    0.9218}
\definecolor{colSolar}{rgb}     {    0.0704,    0.7457,    0.7258}
\definecolor{colWind}{rgb}      {    0.5044,    0.7993,    0.3480}
\definecolor{colBioenergy}{rgb} {    0.9871,    0.7348,    0.2438}
\definecolor{colImported}{rgb}  {    0.9769,    0.9839,    0.0805}

\journal{Applied Energy}
\begin{document}
\begin{frontmatter}

\title{Short-Term Forecasting of Energy Production and Consumption Using Extreme Learning Machine: A Comprehensive \texttt{MIMO} based \texttt{ELM} Approach}

\author[label1]{Cyril Voyant\orcidlink{0000-0003-0242-7377}} \corref{cor1}
\ead{cyril.voyant@mines-psl.eu} 


\author[label2,label3]{Milan Despotovic\orcidlink{0000-0003-3144-5945}}
\ead{milan.z.despotovic@gmail.com} 

\author[label2]{Luis Garcia-Gutierrez\orcidlink{0000-0002-3480-1784}}
\ead{garcia_gl@univ-corse.fr} 

\author[label1,label2]{Mohammed Asloune\orcidlink{0009-0004-0183-2156}}
\ead{asloune_m1@univ-corse.fr}

\author[label1]{Yves-Marie Saint-Drenan\orcidlink{0000-0003-1471-1092}}
\ead{yves-marie.saint-drenan@minesparis.psl.eu}

\author[label2]{Jean-Laurent Duchaud\orcidlink{0000-0001-7490-7260}}
\ead{duchaud_jl@univ-corse.fr}

\author[label2]{Ghjuvan Antone Faggianelli\orcidlink{0000-0002-9954-5910}}
\ead{faggianelli_ga@univ-corse.fr}

\author[label1]{Elena Magliaro}
\ead{elena.magliaro@minesparis.psl.eu}

\affiliation[label1]{organization={Mines Paris, PSL University},
            addressline={Centre for observation, impacts, energy (O.I.E.)}, 
            city={Sophia-Antipolis},
            postcode={06904}, 
            state={côte d'azur},
            country={France}}
\affiliation[label2]{organization={SPE Laboratory, UMR CNRS 6134},
            addressline={University of Corsica Pasquale Paoli}, 
            city={Ajaccio},
            postcode={20000 }, 
            state={Corsica},
            country={France}}
\affiliation[label3]{organization={Faculty of Engineering},
            addressline={University of Kragujevac}, 
            city={Kragujevac},
            postcode={34000}, 
            state={Šumadija},
            country={Serbia}}

\begin{abstract}
A novel methodology for short-term energy forecasting using an Extreme Learning Machine ($\mathtt{ELM}$) is proposed. Using six years of hourly data collected in Corsica (France) from multiple energy sources (solar, wind, hydro, thermal, bioenergy, and imported electricity), our approach predicts both individual energy outputs and total production (\cyr{including imports, which closely follow energy demand, modulo losses)} through a Multi-Input Multi-Output ($\mathtt{MIMO}$) architecture.  
To address non-stationarity and seasonal variability, sliding window techniques and cyclic time encoding are incorporated, enabling dynamic adaptation to fluctuations. The $\mathtt{ELM}$ model significantly outperforms persistence-based forecasting, particularly for solar and thermal energy, achieving an $\mathtt{nRMSE}$ of $17.9\%$ and $5.1\%$, respectively, with $\mathtt{R^2} > 0.98$ (1-hour horizon). The model maintains high accuracy up to five hours ahead, beyond which renewable energy sources become increasingly volatile. While $\mathtt{MIMO}$ provides marginal gains over Single-Input Single-Output ($\mathtt{SISO}$) architectures and offers key advantages over deep learning methods such as $\mathtt{LSTM}$, it provides a closed-form solution with lower computational demands, making it well-suited for real-time applications, including online learning. Beyond predictive accuracy, the proposed methodology is adaptable to various contexts and datasets, as it can be tuned to local constraints such as resource availability, grid characteristics, and market structures. 
\end{abstract}

\begin{graphicalabstract}
\begin{figure}[h!]
\centering
\begin{tikzpicture}[scale=0.8, every node/.style={transform shape, minimum height = 3cm}]
    \tikzset{header/.style={align=center, font=\small, yshift=1em}}
        \node[draw, rounded corners, fill=inputcolor!20, text width=2.8cm, align=center, font=\small] (input) at (0,0) 
    {Energy Production \& Consumption \\ (Solar, Wind, Hydro, etc.)};
    \node[draw, rounded corners, fill=inputcolor!30, text width=3.0cm, align=center, font=\small] (features) at (4,0) 
    {Feature Engineering \\ (Sliding Window, Time Encoding)};
    \node[draw, rounded corners, fill=hiddencolor!20, text width=2.2cm, align=center, font=\small] (hidden) at (8,0) 
    {$\mathtt{ELM}$ \\ Hidden Layer};
    \node[draw, rounded corners, fill=outputcolor!20, text width=3.2cm, align=center, font=\small] (output) at (12,0) 
    {Predicted Energy \\ Outputs};
    \node[draw, rounded corners, fill=evaluationcolor!20, text width=2.5cm, align=center, font=\small] (evaluation) at (16,0) 
    {Evaluation \\ ($\mathtt{RMSE}$, $\mathtt{MAE}$, $\mathtt{R^2}$)};
        \draw[->, thick] (input) -- (features);
    \draw[->, thick] (features) -- (hidden);
    \draw[->, thick] (hidden) -- (output);
    \draw[->, thick] (output) -- (evaluation);    
    \path (input.west) -- (evaluation.east) coordinate[midway](H1);
        \node[align=center, font=\bfseries\small, yshift=-2em] at (H1|-input.south) 
    {Energy Forecasting Using $\mathtt{ELM}$ in Multiple-Input Multiple-Output Context};
        \node[header] at (input.north) {Input: Historical Data};
    \node[header] at (features.north) {Feature Transformation};
    \node[header] at (hidden.north) {$\mathtt{ELM}$ Training};
    \node[header] at (output.north) {Energy Predictions};
    \node[header] at (evaluation.north) {Model Evaluation};
\end{tikzpicture}
\end{figure}
\end{graphicalabstract}

\begin{highlights}
    \item[\(\checkmark\)] A novel Extreme Learning Machine-based Multi-Input Multi-Output ($\mathtt{MIMO}$) framework for short-term energy forecasting is proposed.
    \item[\(\checkmark\)] The model accurately predicts energy outputs from multiple renewable and non-renewable sources, achieving high accuracy up to five hours ahead.
    \item[\(\checkmark\)] Performance is evaluated using normalized $\mathtt{RMSE}$, $\mathtt{MAE}$, and $\mathtt{R^2}$, demonstrating significant improvements over the persistence $\mathtt{SISO}$ model and deep learning-based $\mathtt{LSTM}$ approaches.
    \item[\(\checkmark\)] Accurate forecasting can be a powerful decision-support tool, optimizing dispatch and aiding renewable integration within operational constraints.
\end{highlights}


\end{frontmatter}
\section{Introduction}
The integration of renewable energy ($\mathtt{RE}$) sources, such as solar and wind, is important for reducing greenhouse gas emissions and mitigating climate change \cite{moretti2020grid, basit2020limitations, POMMERET2022104273}. However, their weather-dependency introduces variability in power supply, posing challenges for grids originally designed for stable fossil fuel or nuclear generation \cite{notton2018intermittent}. Ensuring grid stability and balancing supply and demand remain key challenges in high-$\mathtt{RE}$ systems \cite{moretti2020grid}, making accurate forecasting essential for grid management and energy storage optimization \cite{su15097087}. Despite the inherent variability of wind and solar \cite{TALARI20181953}, machine learning advances now enable more precise short-term energy predictions through large datasets and advanced computational methods \cite{ALAMO2019111}.

\subsection{Motivation for Short-Term Forecasting in Multi-Source Energy Systems}
The increasing complexity of multi-source grids necessitates real-time adjustments to balance supply and demand, minimize fossil fuel reliance, and improve grid stability.  Effective decision-making requires not only accurate predictions but also a broader assessment of resource availability, generation constraints, and transmission delays. Forecasting should be viewed as a decision-support tool for system operators, integrated into broader optimization frameworks that consider real-time constraints and flexibility mechanisms.

The increasing role of self-consumption in distributed photovoltaic ($\mathtt{PV}$) systems further complicates demand forecasting. As more energy is consumed directly at the production site, $\mathtt{PV}$ capacity translates less directly into injected power, instead reducing the total power drawn from the grid. This shift necessitates integrated forecasting tools that account for both generation and consumption, a need that will only intensify with rising self-consumption levels.

Traditional statistical methods \cite{voyant2017machine} often struggle with the nonlinear interactions within complex multi-source energy systems. Machine learning techniques, such as Extreme Learning Machines ($\mathtt{ELM}$), offer improved accuracy by integrating meteorological data, historical trends, and grid conditions \cite{teo2015forecasting, al2018extreme, la2021new}. A Multiple-Input Multiple-Output ($\mathtt{MIMO}$) approach extends this capability by simultaneously predicting multiple energy variables (e.g., solar irradiance, wind speed, demand, and grid load). This holistic approach enables better coordination of generation, storage, and dispatch, leading to more efficient and reliable grid operation by capturing the complex interdependencies between these variables.

While thermal power output is influenced by operational decisions, fuel availability, and technical constraints, forecasting its potential evolution remains valuable, especially when combined with information on hydropower and storage. This allows for better anticipation of critical periods and cost fluctuations. Transmission system operators also rely on accurate forecasts to anticipate power flows. Inaccurate predictions can lead to costly imbalances, impacting network stability, a key concern for these operators. 

\cyr{Forecasts can also serve as a substitute for real-time measurements in the (many) cases where such data are unavailable}. Beyond operational concerns, the increasing digitalization of energy systems introduces cybersecurity risks. False data injection attacks and manipulation of measurement signals highlight the vulnerability of grid management systems \cite{he2016cyber, reda2022comprehensive}. Comparing predicted values with actual grid measurements can help detect anomalies, improving both operational efficiency and security \cite{le2022smart, fuller2020digital}.

As a result, while $\mathtt{MIMO}$ forecasting is one component of effective grid management, its ability to integrate diverse data sources, account for system complexities, and support real-time decision-making makes it an important tool for navigating the technical, economic, and increasingly decentralized challenges of the evolving energy landscape. 

This includes not only traditional grid operations but also the efficient management of distributed resources like microgrids, where forecasting plays a vital role in optimizing energy use and maximizing the benefits of local generation and storage. Specifically, for interconnected microgrids, forecasting power limitations and electricity prices on the main grid is challenging for optimizing energy trading. A forecasting horizon of several hours enables better management of storage, load, and potentially even the scheduling of generation resources, allowing microgrids to maximize economic benefits or minimize their environmental impact. 

\cyr{Our hypothesis in developing the $\mathtt{MIMO}$ approach is that some of the errors in the predictions of individual energy sources can balance each other out. From a statistical point of view, adopting a holistic approach where each energy source and their sum are predicted by the same algorithm allows these balancing effects to be taken into account during the learning process, potentially resulting in a better forecast of residual demand compared to a traditional ($\mathtt{SISO}$) approach. Another possible strategy is reconciliation: when several quantities are linked by linear constraints, one can 'reconcile' univariate forecasts so that they satisfy those constraints, for example through a well-chosen projection.}

The structure of this paper is as follows: Section \ref{sec:relatedworks} reviews previous researches on short-term energy forecasting, focusing on machine learning approaches, in particular the use of $\mathtt{ELM}$ in energy systems. Based on the energy problem and the relevant ad-hoc literature, we introduce Section~\ref{obj} and outline the objectives of this study. The section \ref{sec:methodologie} presents the methodology, detailing the $\mathtt{ELM}$-based forecasting model, preprocessing, architecture and performance measures. Section \ref{sec:Results} describes the experimental setup, including dataset, evaluation framework and reference models, followed by a presentation of the results, discussing the accuracy, computational efficiency and implications of real-time forecasting in multi-source energy networks, illustrated by Corsica. Section \ref{Sec:Conclusion} concludes by summarizing the main results, addressing limitations and suggesting directions for future research.

\section{Related Works}
\label{sec:relatedworks}
This section reviews the main advancements in $\mathtt{RE}$ forecasting. We discuss various methodologies, underscoring their merits and drawbacks, while emphasizing the benefits of $\mathtt{ELM}$ over conventional techniques.

\subsection{Energy Forecasting Approaches}
Energy forecasting plays an important role in balancing supply and demand in modern electricity systems, especially with the increasing integration of $\mathtt{RE}$ sources. Recent advances in machine learning and, in general, the development of artificial intelligence ($\mathtt{AI}$) have revolutionized the ability to accurately predict $\mathtt{RE}$ and non-$\mathtt{RE}$ production. \cite{bishaw2024, benti2023, alansari2023}. Techniques such as Support Vector Regression ($\mathtt{SVR}$), deep learning models, and hybrid AI approaches have shown great potential in capturing complex patterns in energy datasets. In \cite{elsaraiti2024shortterm} neural networks are used to capture seasonal patterns and abrupt variations in energy consumption data, which is especially useful for managing intermittent energy sources that exhibit high variability.

In recent research, Saglam et al. (2023) \cite{Saglam2023} introduced a new method for predicting electricity demand in Turkey. Authors used Particle Swarm Optimization ($\mathtt{PSO}$) with Support Vector Regression ($\mathtt{SVR}$). This hybrid $\mathtt{PSO-SVR}$ model demonstrated suitable improvements, illustrating how optimization techniques can enhance machine learning applications in energy forecasting. Along similar lines, Ardabili et al. (2022) \cite{Ardabili2022} conducted an in-depth review of deep learning applications for forecasting energy demand. Their work emphasizes the advantages of multimodal $\mathtt{AI}$ systems in managing both $\mathtt{RE}$ and non-$\mathtt{RE}$ sources, offering insights into the evolving role of $\mathtt{AI}$ in energy systems management.

In addition, Motwakel et al. (2023) \cite{csse.2023.037735} proposed a deep learning-based model that combines $\mathtt{SVR}$, $\mathtt{Lasso}$ and $\mathtt{Ridge}$ regression to predict non-$\mathtt{RE}$ consumption, achieving state-of-the-art accuracy in multimodal forecasting scenarios. 
On the $\mathtt{RE}$ side, Ding et al. (2022) \cite{Ding2022} integrated long-term memory models ($\mathtt{LSTM}$) and $\mathtt{SVR}$ to improve solar power generation forecasting, demonstrating the effectiveness of hybrid models in reducing prediction errors. In addition, Revathi (2023) \cite{Revathi2023} conducted a wide-ranging investigation of AI-driven techniques for $\mathtt{RE}$ production, focusing on deep learning architectures, highlighting their potential for optimizing both production and demand management.

\subsection{Advantages of \texttt{ELM} in Energy Forecasting}
$\mathtt{ELM}$ shows important potential in energy demand and sources forecasting, particularly when applied to $\mathtt{RE}$ sources. The rapid learning capacity and generalization performance of $\mathtt{ELM}$ have been widely recognized in recent studies. Abd El-Aziz et al. (2022) \cite{ABDELAZIZ20229447} demonstrated that $\mathtt{ELM}$ improves the accuracy of consumption forecasting for $\mathtt{RE}$ sources by integrating deep learning models for hybrid $\mathtt{RE}$ systems. Recently, Li et al. (2022) \cite{en11092475} also took up the challenge by using tools like $\mathtt{ELM}$ and $\mathtt{SVR}$ to understand and predict the carbon emissions. They found that $\mathtt{ELM}$, a model simple in spirit but powerful in scope, had an advantage. Especially in rugged places where coal and oil still have power, $\mathtt{ELM}$ proved relevant.

The same year, Sibtain et al. (2022) \cite{SIBTAIN2022115703} joined the topic. They designed a multi-stage deep learning machine, a powerful ally for predicting wind vagaries. For those working on $\mathtt{RE}$, knowing the next value of the wind speed can be complicated or even impossible. 
Their results highlighted a truth: $\mathtt{ELM}$ has, time and again, outperformed old methods, proving itself in the wild terrain of green energy and fossil fuel dependency.

$\mathtt{ELM}$ allows us to imagine a future in which energy demands and resources are easily predictable, however complex the challenge may be. 
To our knowledge, no studies or tools have been proposed in the literature for the simultaneous forecasting of all production sources and electricity consumption within a region. This study positions itself as a pioneer in this field.

\section{Objective of the Study}
\label{obj}
The aim of this work is to propose a streamlined and efficient methodology for short-term energy forecasting without relying on data stationarity. Energy systems, including both $\mathtt{RE}$ and non-$\mathtt{RE}$ sources as well as electricity demand, exhibit non-stationary behavior due to factors such as seasonal variations, maintenance schedules, and operational fluctuations. However, traditional forecasting methods often assume stationary processes \cite{Han2024}, which does not always hold in energy management where cyclic patterns and dynamic trends play an essential role.
To address this challenge, our approach integrates temporal cues and periodic patterns directly into its structure, allowing the model to adapt dynamically without requiring stationarization. This preserves the intrinsic characteristics of the data while improving predictive accuracy. \cyr{An alternative strategy would be to build a separate model for each hour of the day, but our method captures these variations within a unified framework} 

Moreover, this study investigates the impact of Single-Input Single-Output ($\mathtt{SISO}$) vs. Multiple-Input Multiple-Output ($\mathtt{MIMO}$) approaches in forecasting energy sources and demand, evaluating their respective advantages and limitations. In addition, a comparison between a simple  Extreme Learning Machine ($\mathtt{ELM}$)-based model against a more complex one, computationally intensive, and less interpretable denoted Long Short-Term Memory ($\mathtt{LSTM}$)-based approach. While $\mathtt{ELM}$ offers fast learning and precise predictions with minimal computational requirements, $\mathtt{LSTM}$ architectures provide advanced feature extraction and long-term dependency modeling at the cost of increased complexity. \cyr{A simpler alternative (with a trade-off in accuracy) is the $\mathtt{TSMixer}$ architecture, a feedforward model designed for time series forecasting, which yielded good results on net consumption forecasting. Strong performance was also obtained with Temporal Fusion Transformers ($\mathtt{TFT}$), which are attention-based models combining recurrent and interpretable components, though they are more complex.}

To assess performance, the models are evaluated using Root Mean Square Error ($\mathtt{RMSE}$), Mean Absolute Error ($\mathtt{MAE}$), and the coefficient of determination $ \mathtt{R^2} $. Determining whether the proposed framework outperforms traditional approaches and providing insights into the trade-offs between simplicity, accuracy, and interpretability in energy forecasting.
\section{Materials \& Methods}
\label{sec:methodologie}
In this study, time series representing the hourly electric power produced by different sources are collected and analyzed to assess their respective contributions to Corsica’s overall energy generation. The workflow of the Multiple Input, Multiple Output ($\mathtt{MIMO}$) predictive method is detailed in \ref{annexA}. In machine learning, Single Input, Single Output ($\mathtt{SISO}$) and  $\mathtt{MIMO}$ are two modeling approaches with distinct trade-offs. $\mathtt{SISO}$ $\mathtt{ELM}$ trains separate models for each output variable using a single input feature, offering better interpretability and flexibility but increasing computational cost. In contrast, $\mathtt{MIMO}$ $\mathtt{ELM}$ predicts all outputs simultaneously using multiple inputs, capturing interdependencies between outputs and being more computationally efficient, though potentially sacrificing accuracy for individual predictions. $\mathtt{SISO}$ is usually ideal when outputs are independent, while $\mathtt{MIMO}$ is preferable when outputs are correlated, improving overall consistency and reducing training time.

\subsection{Data Description}
While such a study could be performed using data from any region, insular data were specifically chosen to address the unique challenges of islands, such as isolated grids, supply-demand imbalances, and seasonal variations. 

Limited interconnections and high $\mathtt{RE}$ intermittency further complicate grid stability \cite{NOTTON20191157}. Small grids with low inertia lead to frequency variability and voltage instability. Limited spatial aggregation results in fluctuating $\mathtt{RE}$ generation, sometimes exceeding safety thresholds, requiring curtailments. Seasonal demand variations necessitate flexible power generation, with Corsican electricity demand ranging from 130 MW to 450 MW depending on the period. 
Corsica, the smallest and most mountainous island in the western Mediterranean (8,680 km$^2$, 568 m avg. altitude), had approximately 350,000 inhabitants in 2022 \cite{insee_essentiel_corse_2024}. Tourism, with 3 million annual visitors mainly from May to September, significantly impacts electricity demand. Currently, Corsica electrical system has a total installed capacity around 1 GW, including 369 MW from thermal sources, 225 MW from hydropower, 233 $MW_{p}$ from solar PV, 18 MW from wind, 2 MW from biogas, 150 MW from imports, and 6 MW from battery storage. The island is partially connected to Italy through the $\mathtt{SACOI}$ interconnection (50 MW $\mathtt{DC}$, operational since 1986) and $\mathtt{SARCO}$ (100 MW $\mathtt{AC}$, since 2006). However, Corsica can only import electricity due to commercial constraints.
Electricity production remains highly dependent on diesel generators, resulting in CO$_2$ emissions up to 15 times higher than those in mainland France. Around 70\% of electricity is still fossil-fuel-based \cite{NOTTON20191157,cre_transition_energetique}.

The dataset used in this study includes electricity demand and key energy sources such as solar, wind, hydropower, thermal, bioenergy, and imports, covering the six previous years (2018-2023) at hourly resolution, resulting in 368,256 data points, with additional data for leap years. These data, managed by Electricité de France ($\mathtt{EDF}$), the leading electricity provider in France and responsible for grid management in Corsica, are publicly available at \url{https://www.data.corsica/pages/portail/} \cite{edf_open_data_corse}.
The dataset is \cyr{based on measurements aggregated as average values over hourly time slots (hourly granularity)}, enabling detailed temporal analysis. \cyr{While hourly data may be considered coarse compared to the 15-minute resolution adopted by spot markets, this level of temporal granularity remains sufficient to identify production peaks and observe differences in behavior across energy sources}. For data processing, several transformations were applied to prepare the time series data for forecasting and deeper analysis. A sliding window approach was implemented to capture temporal dependencies, where each observation consists of a sequence of values within a window of previous time steps, $X_i = [x_{i}, x_{i+1}, \ldots, x_{i+W-1 }]$. This method enhances the model’s ability to learn patterns over time.

\subsection{Feature Engineering}
Sine and cosine coding \cite{GAIRAA2022890} were applied to capture cyclical temporal features using two harmonics defining by: $\text{sin}_T(t) = \sin\left(2 \pi \frac{t}{T}\right), \quad \text{cos}_T(t) = \cos\left(2 \pi \frac{t}{T}\right)$ ; where $t$ represents the hour of the day (0 to 23) and $T$ is 24 for the daily cycle. This preserves the cyclical nature  for the model.
Concerning missing data and cleaning process, often due to monitoring or $\mathtt{IT}$ issues, a linear interpolation was operated filling the gaps with the average of neighboring values. The longest data gap lasted only a few hours; therefore, the gap-filling process does not significantly alter the dataset.
 With these preprocessing steps, including sliding windows, temporal encoding, and robust missing data handling, the dataset is ready for simulations and to propose robust conclusions. 
To quantify the link between energy variables, we suggest the Mutual Information ($\mathtt{MI}$; \cite{10.1007/978-3-642-04138-9_30}), which measures the dependency between two variables, capturing both linear and non-linear relationships, unlike the Pearson correlation, which only detects linear associations. It is defined as:
\begin{equation}
    \mathtt{MI}(X; Y) = \int \int p(x,y) \log \frac{p(x,y)}{p(x)p(y)} dx dy,
\end{equation}
where $p(x,y)$ is the joint probability density, and $p(x), p(y)$ are the marginal distributions. \cyr{$\mathtt{MI}$ can also be interpreted as the Kullback-Leibler divergence between the joint distribution and the product of the marginals, that is, a measure of how far the variables are from being statistically independent.} $\mathtt{MI}$ is always non-negative, bounded by $0 \leq \mathtt{MI}(X; Y) \leq \min(\mathtt{H}(X), \mathtt{H}(Y)),$ where $\mathtt{H}(X)$ and $\mathtt{H}(Y)$ are the entropies of $X$ and $Y$. To ensure comparability, we normalize $\mathtt{MI}$:
\begin{equation}
    \mathtt{MI}_{\text{norm}}(X; Y) = \frac{\mathtt{MI}(X; Y)}{\sqrt{\mathtt{H}(X)\mathtt{H}(Y)}}, \quad 0 \leq \mathtt{MI}_{\text{norm}}(X; Y) \leq 1.
\end{equation}
The analysis highlights strong dependencies: thermal and bioenergy ($0.535$), hydropower and thermal ($0.373$), and solar with bioenergy ($0.468$). Total production is mainly influenced by thermal ($0.256$), hydro ($0.129$), imports ($0.231$), and solar ($0.124$). These relationships hint for a structured interaction between energy sources, reinforcing the relevance of a Multi-Input Multi-Output ($\mathtt{MIMO}$) approach for improved forecasting. Typically, a normalized Mutual Information ($\mathtt{MI}_{\text{norm}}$) above 0.2-0.3 indicates a notable relationship, while values exceeding 0.5 denote strong dependence.

\subsection{\texttt{ELM} in Prediction Context}
The Extreme Learning Machine ($\mathtt{ELM}$) is a type of feedforward neural network, that is simple and very effective. Unlike traditional networks that loop and keep sending information from one side to the other through layers of complexity (backpropagation), $\mathtt{ELM}$ stands out for its simplicity. It is a direct feedback neural network, with fast learning: a single step, no unnecessary movements, no loss of precision. While a conventional $\mathtt{MLP}$ (multilayer perceptron) network is theoretically less efficient than an $\mathtt{ELM}$, a deep learning network (such as $\mathtt{LSTM}$; see appendix \ref{annexB}) will be more optimal than an $\mathtt{ELM}$, but will also be much harder to train. For the $\mathtt{ELM}$, a few initializations and linear regressions suffice, making it eligible for “continuous” training over a sliding window. $w$ is chosen using a brute-force method that tests all possible configurations ($w=24:96$ in 24h steps), keeping the number of hidden nodes constant ($2^{12}$). In this context, $w=48$ remains the best option.

$\mathtt{ELM}$ is composed of three parts: an input layer, a hidden layer and an output layer. The network only needs $I$ input features to becomes operational. The hidden layer contains $H$ neurons, their number being selected according to the task to be accomplished; it is advisable to propose enough to accomplish the work, to match the weight of the task. From there, $O$ output neurons await the result, ready to compute predictions, which we hope are robust and effective. Below the main parameters of the $\mathtt{ELM}$ forecasting tool using the Mean Squared Error ($\mathbf{MSE}$) minimizing across all output variables without additional weighting:
\cyr{
\begin{itemize}
    \item[\(\checkmark\)]  \textbf{Input Layer}: The input layer receives the feature vector $\mathbf{x} = [x_1, x_2, \ldots, x_n]$;
    \item[\(\checkmark\)]  \textbf{Hidden Layer}: The hidden layer applies a linear combination of inputs followed by a $\mathtt{ReLU}$ activation function $h_j(\mathbf{x}) = \max(0, \mathbf{w}_j^T \mathbf{x} + b_j)$, where $\mathbf{w}_j$ is the weight vector and $b_j$ the bias for the $j^{th}$ hidden neuron. The hidden layer output is $\mathcal{H} = [h_1(\mathbf{x}), \ldots, h_H(\mathbf{x})]^T$ ($H$ is the number of hidden nodes);
    \item[\(\checkmark\)]  \textbf{Output Layer}: The final prediction is a linear combination of hidden layer outputs $\mathbf{y} = \mathcal{H}^T \mathcal{W}$, where $\mathcal{W}$ is the matrix of output weights;
    \item[\(\checkmark\)]  \textbf{Training Process}: $\mathtt{ELM}$ training consists of two phases: random weight initialization between the input and hidden layers, and output weight calculation $\mathcal{W} = \mathcal{H}^+ \mathbf{Y}$, where $\mathbf{Y}$ is the target output and $\mathcal{H}^+$ is the Moore-Penrose pseudo-inverse of $\mathcal{H}$;
    \item[\(\checkmark\)]  \textbf{Multiple Initializations}: To ensure robustness, multiple random weight initializations are performed (50 in the case of this study). The best configuration, determined by evaluating a performance metric such as $\mathtt{RMSE}$ ($= \frac{1}{N} \sum_{i=1}^{N} (y_i - \bar{y})^2$) computed in-sample, is selected.
\end{itemize}}
In \ref{annexB}, the schema of the multivariate $\mathtt{ELM}$ used in the simulations is presented, with $7 \times 48+2$ inputs and 7 outputs. The inputs correspond to all electrical variables (7), spanning 2 days and two time indices $\text{sin}_T(t)$ and $\text{cos}_T(t)$, while the outputs represent each of the 7 electrical variables at the forecast horizon. 
For validation, a persistence model, a simple and reliable reference, is used. It is the measure against which the predictions of the $\mathtt{ELM}$ model are compared. In the absence of complex patterns, it intervenes, assuming that the future closely resembles the last known moment. Known by some as naive forecasting, it considers each subsequent step as an echo of the previous one.
Mathematically, the forecast, $\hat{y}_{t+h}$, for a future point $t+h$, is defined by the relation $\hat{y}_{t+h} = y_t$, where $y_t$ marks the last value recorded at time $t$, and $h$ is the deviation that we project into the future. Its simple logic proves particularly effective when past numbers have an influence on what will follow.
When it comes to predicting the energy mix, this is often our only option because other models based on more in-depth knowledge have not yet surfaced (unstable installed power over time, breakdowns, maintenance, variability, seasonality, etc.). 

\subsection{Evaluation Metrics}
To evaluate the performance of the $\mathtt{ELM}$ forecasting model, we use key error metrics: $\mathtt{RMSE}$, $\mathtt{MAE}$, and $\mathtt{R^2}$, along with their normalized versions. These metrics provide a comprehensive assessment of accuracy and facilitate comparisons with benchmark models.
$\mathtt{RMSE}$ measures prediction errors by emphasizing larger deviations, making it sensitive to outliers. 
$\mathtt{MAE}$, defined as $\mathtt{MAE} = \frac{1}{N} \sum_{i=1}^{N} |y_i - \bar{y}|,$ captures the average absolute error and treats all deviations equally, offering robustness against extreme values. 
$\mathtt{R^2}$ quantifies the variance and ranges from $0$ (no explained variance) to $1$ (perfect fit). 
To ensure comparability across datasets, we use normalized error metrics:
\begin{equation}
    \mathtt{nRMSE} = \frac{\mathtt{RMSE}}{\bar{y}}, \quad \mathtt{nMAE} = \frac{\mathtt{MAE}}{\bar{y}}.
\end{equation}
Performance is benchmarked against a persistence model using the gain metric (\cyr{or skill score}):
\begin{equation}
    \mathtt{G} = \frac{\mathtt{RMSE_{persistence}} - \mathtt{RMSE_{ELM}}}{\mathtt{RMSE_{persistence}}},
\end{equation}
where a positive $\mathtt{G}$ indicates superior $\mathtt{ELM}$ performance. For instance, if the persistence model yields $\mathtt{RMSE} = 0.50$ and $\mathtt{ELM}$ achieves $0.45$, then $\mathtt{G} = 0.1$, reflecting a 10\% error reduction.
\section{Results and Discussion}
\label{sec:Results}
All codes and datasets used in this study are publicly available at the following repository: \url{https://github.com/cyrilvoyant/Data_Viz_2024}. For the simulations, 80\% of the data were used for in-sample training, while the remaining 20\% were allocated for out-of-sample testing (error metrics were calculated exclusively on the data not involved in the learning phase). The separation between samples sets is chronological: the first values are used for training, and the subsequent ones for testing (hindcast process).
In the following, we will refer to “electricity consumption” as the total amount produced by all energy sources, disregarding losses (even though in some cases, these may reach a few percent).

\subsection{Feasibility}
\begin{figure}
\centering
\includegraphics[width=0.7\textwidth]{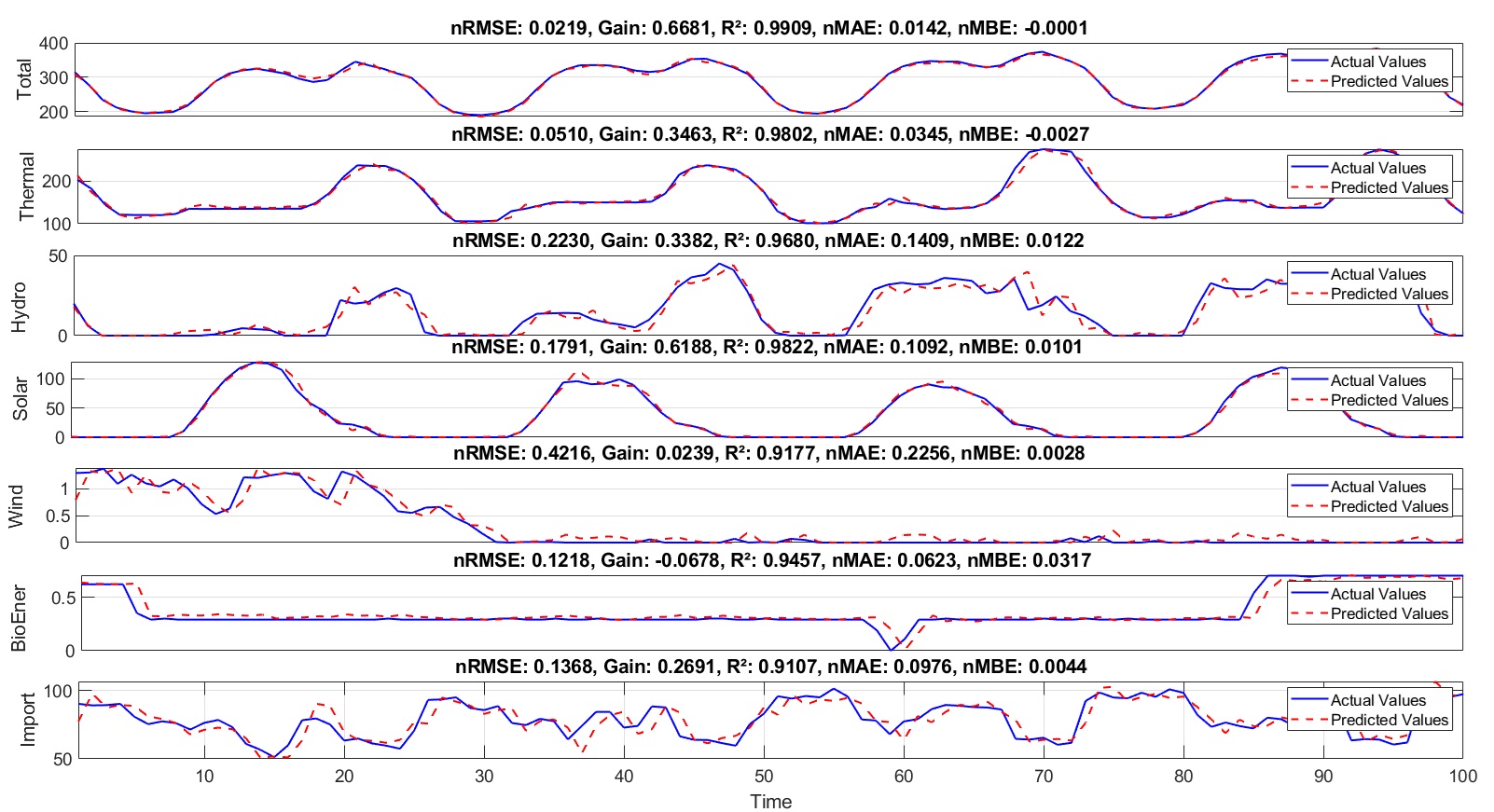}
\caption{Profile of the 7 Predictions (1h horizon) for $\sim$ 4 Consecutive Days (100h) in Winter. Electrical Variables in MW and time in hour.}
\label{fig:profile}
\end{figure}

In Figure \ref{fig:profile}, the prediction results for the six energy sources and the total electrical consumption of Corsica are presented. Error metrics are computed over 1 year of hourly predictions for forecasting horizon ranging from 1 to 10 hours. These curves confirm that the forecasting method is valid, both visually and in terms of the error metrics for each plot, demonstrating its efficiency. For instance, on the first day (between 8:00 and 22:00), the two peaks in Total production $\approx$ Total consumption (morning and evening) are attributed to solar and thermal energy, respectively. The cross-pattern of the method enhances its responsiveness. 
The contribution of bio-energy to the energy mix is minimal, with low variability, suggesting that a simple persistence model is more efficient due to the negative gain value. 
Wind energy, on the other hand, remains challenging to predict, with an $\mathtt{nRMSE}$ equal to 0.4216. Keeping in mind that the wind turbine installed power is only 18 MW spread over three wind farms relatively close geographically. 
For solar energy, the error metrics are particularly favorable; however, it is important to note that the spatial distribution of solar stations reduces resource variability, making the prediction easier. 
Across all horizons, the conclusions remain consistent, indicating the proposed methodology is efficient. Some examples of prediction using the $\mathtt{ELM}$ methodology are available in \ref{annexC} concerning horizons $h+5$ and $h+6$.

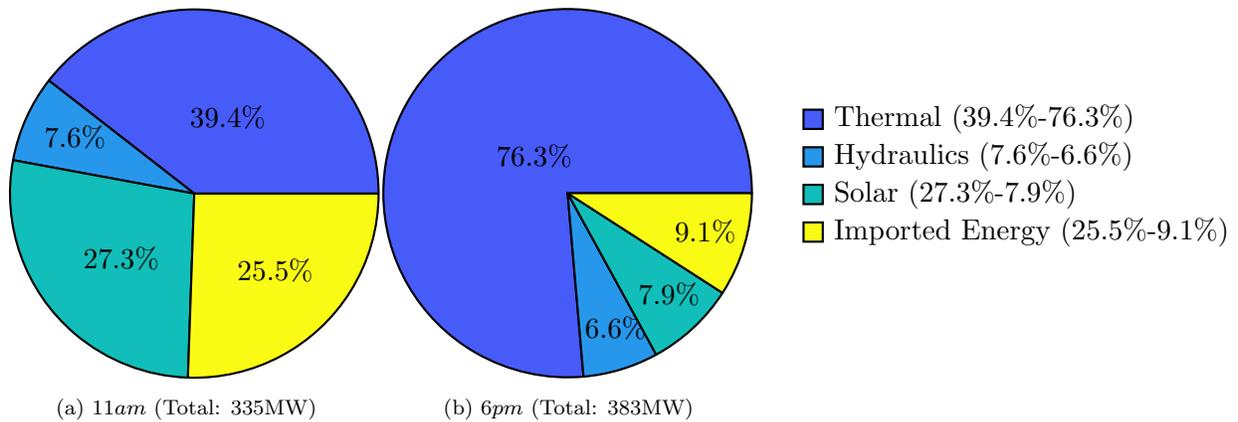
\begin{figure}
    \begin{subfigure}{0.3\linewidth}
    \centering
    \begin{tikzpicture}
        \pie[sum=auto, text=pin, text=legend, after number=\%, radius=2.45, 
             color={colThermal, colHydraulic, colSolar, colImported}, 
             before number=\phantom{0}]{
             39.4/Thermal,
             7.6/Hydraulics,
             27.3/Solar,
             25.5/Imported Energy}
    \end{tikzpicture}
    \caption{$11am$ (Total: $335$MW)}
    \end{subfigure}
	\begin{subfigure}{0.3\linewidth}
	\centering
     \begin{tikzpicture}
        \pie[sum=auto, text=pin, text=legend, after number=\%, radius=2.45, 
             color={colThermal, colHydraulic, colSolar, colImported}, 
             before number=\phantom{0}]{
             76.3/Thermal (39.4\%-76.3\%),
             6.6/Hydraulics (7.6\%-6.6\%),
             7.9/Solar (27.3\%-7.9\%),
             9.1/Imported Energy (25.5\%-9.1\%)}
    \end{tikzpicture}
   	\caption{$6pm$ (Total: $383$MW)}
	\end{subfigure}
	\caption{Breakdown of Energy Consumption in Percentages. Wind and bioenergy not represented ($\le 0.1 \%$)}
 	\label{fig:energy_consumption}
\end{figure}

The distribution of energy sources at $11am$ and $6pm$ on the third days (Figure \ref{fig:profile}) is depicted in Figure \ref{fig:energy_consumption}. In the evening (when the solar resource is tending to disappear), the distribution is predominantly composed of thermal energy, though this pattern is not consistent across different hours (as observed, for instance, during the first peak in demand where Solar is more important; 
This highlights the complexity in forecasting the energy mix and underscores the intricate relationship between time and the various factors captured by the $\mathtt{ELM}$ model, which are neither straightforward nor understandable by the human mind.

\subsection{Forecasting Performance} 
In the following sections, Tables \ref{tab:R2} to \ref{tab:Gain} present the metrics obtained, and Figure \ref{fig:comp} plots them for easier visualization. 
The results presented in Table \ref{tab:R2} clearly illustrate the fact that $\mathtt{R^2}$  values obtained for various energy sources are correlated with the different forecasting horizons ($h$). This observation suggests that the $\mathtt{ELM}$ model not only effectively captures the inherent patterns in energy production data but also consistently does so over time. Indeed, with $\mathtt{R^2}$  values close to $1$, it becomes evident that the $\mathtt{ELM}$ model accounts for a substantial part of the observed variance of dataset, surpassing traditional persistence methods. In light of these results, it appears that the $\mathtt{ELM}$ model is a particularly robust approach for energy output forecasting, offering a nuanced understanding of temporal dependencies while significantly reducing the risk of overfitting, as it is often the case with more complex models.
\begin{small}
\begin{longtable}[t]{@{}cccccccc@{}} 
    \caption{$\mathtt{R^2}$ for Energy Sources (Unitless) computed for 1 year of predictions (out-sample)} \\ 
    \toprule
    Horizon ($h$) & Total & Thermal & Hydraulics & Solar & Wind & Bioenergy & Imported \\ 
    \midrule
    \endfirsthead
    \bottomrule
    \endfoot
    1 & 0.9909 & 0.9802 & 0.9680 & 0.9822 & 0.9177 & 0.9457 & 0.9107 \\
    2 & 0.9800 & 0.9528 & 0.9303 & 0.9493 & 0.8155 & 0.8562 & 0.8021 \\
    3 & 0.9710 & 0.9309 & 0.9073 & 0.9227 & 0.7285 & 0.7635 & 0.7298 \\
    4 & 0.9646 & 0.9139 & 0.8933 & 0.9033 & 0.6548 & 0.6715 & 0.6822 \\
    5 & 0.9602 & 0.8999 & 0.8847 & 0.8910 & 0.5915 & 0.5822 & 0.6539 \\
    6 & 0.9568 & 0.8887 & 0.8786 & 0.8833 & 0.5388 & 0.4971 & 0.6365 \\
    7 & 0.9544 & 0.8800 & 0.8741 & 0.8794 & 0.4884 & 0.4165 & 0.6256 \\
    8 & 0.9527 & 0.8732 & 0.8704 & 0.8776 & 0.4452 & 0.3402 & 0.6186 \\
    9 & 0.9512 & 0.8677 & 0.8676 & 0.8767 & 0.4066 & 0.2687 & 0.6138 \\
    10 & 0.9502 & 0.8626 & 0.8651 & 0.8764 & 0.3746 & 0.2019 & 0.6096 
    \label{tab:R2}
\end{longtable}
\end{small}
Table \ref{tab:nMAE} presents the $\mathtt{nMAE}$ values calculated for different energy sources, thereby providing insights into the prediction accuracy achieved by the $\mathtt{ELM}$ model. These low $\mathtt{nMAE}$ values, notably for thermal and hydraulic energy, imply that the $\mathtt{ELM}$ model’s predictions are quite close to observed values, underscoring its strong predictive performance for these conventional sources. However, in contrast, we observe higher $\mathtt{nMAE}$ values for solar and wind energy. This indicates significant challenges due to the intrinsic variability of these $\mathtt{RE}$ sources. Furthermore, bioenergy and imported energy sources exhibit significant variability, underlining the need for improved forecasting models.
In summary, although conventional energy sources benefit from relatively reliable forecasts, predictions for $\mathtt{RE}$ sources must be refined to better support grid integration and enhance overall stability.
\begin{small}
\begin{longtable}[t]{@{}cccccccc@{}} 
    \caption{$\mathtt{nMAE}$ for Energy Sources (Unitless) computed for 1 year of predictions (out-sample)} \\ 
    \toprule
    Horizon ($h$) & Total & Thermal & Hydraulics & Solar & Wind & Bioenergy & Imported \\ 
    \midrule
    \endfirsthead
    \bottomrule
    \endfoot
    1 & 0.0142 & 0.0345 & 0.1409 & 0.1092 & 0.2256 & 0.0623 & 0.0976 \\
    2 & 0.0221 & 0.0570 & 0.2124 & 0.1838 & 0.3562 & 0.1184 & 0.1494 \\
    3 & 0.0271 & 0.0708 & 0.2490 & 0.2229 & 0.4514 & 0.1644 & 0.1764 \\
    4 & 0.0303 & 0.0799 & 0.2688 & 0.2461 & 0.5269 & 0.2028 & 0.1918 \\
    5 & 0.0323 & 0.0867 & 0.2803 & 0.2570 & 0.5877 & 0.2356 & 0.2003 \\
    6 & 0.0337 & 0.0917 & 0.2882 & 0.2626 & 0.6373 & 0.2639 & 0.2050 \\
    7 & 0.0347 & 0.0954 & 0.2932 & 0.2636 & 0.6827 & 0.2892 & 0.2076 \\
    8 & 0.0355 & 0.0980 & 0.2975 & 0.2616 & 0.7215 & 0.3118 & 0.2092 \\
    9 & 0.0360 & 0.1003 & 0.3008 & 0.2610 & 0.7548 & 0.3319 & 0.2102 \\
    10 & 0.0365 & 0.1024 & 0.3032 & 0.2610 & 0.7829 & 0.3500 & 0.2113 
    \label{tab:nMAE}
\end{longtable}
\end{small}
The results in Table \ref{tab:nMBE} show $\mathtt{nMBE}$  values close to zero for all energy sources (except for large horizons concerning Bio-energy), signifying that the $\mathtt{ELM}$ model provides unbiased forecasts. This is particularly important in energy forecasting, where systematic overestimation or underestimation can lead to significant operational and financial consequences. As a remark, thermal production and consumption are consistently underestimated, while hydraulics, solar, and imports are always overestimated.
\begin{small}
\begin{longtable}[t]{@{}cccccccc@{}} %
    \caption{$\mathtt{nMBE}$ for Energy Sources (Unitless) computed for 1 year of predictions (out-sample)} \\ 
    \toprule
    Horizon ($h$) & Total & Thermal & Hydraulics & Solar & Wind & Bioenergy & Imported \\ 
    \midrule
    \endfirsthead
    \bottomrule
    \endfoot
    1 & -0.0001 & -0.0027 & 0.0122 & 0.0101 & 0.0028 & 0.0317 & 0.0044 \\
    2 & -0.0002 & -0.0062 & 0.0206 & 0.0121 & 0.0026 & 0.0668 & 0.0096 \\
    3 & -0.0003 & -0.0092 & 0.0258 & 0.0138 & 0.0024 & 0.0972 & 0.0140 \\
    4 & -0.0004 & -0.0119 & 0.0290 & 0.0147 & 0.0019 & 0.1243 & 0.0177 \\
    5 & -0.0004 & -0.0144 & 0.0315 & 0.0133 & 0.0011 & 0.1483 & 0.0210 \\
    6 & -0.0005 & -0.0166 & 0.0334 & 0.0113 & 0.0000 & 0.1695 & 0.0237 \\
    7 & -0.0006 & -0.0186 & 0.0344 & 0.0090 & -0.0010 & 0.1884 & 0.0260 \\
    8 & -0.0006 & -0.0203 & 0.0348 & 0.0071 & -0.0017 & 0.2057 & 0.0280 \\
    9 & -0.0007 & -0.0217 & 0.0351 & 0.0050 & -0.0029 & 0.2213 & 0.0297 \\
    10 & -0.0008 & -0.0230 & 0.0356 & 0.0038 & -0.0039 & 0.2354 & 0.0311 
    \label{tab:nMBE}
\end{longtable}
\end{small}
In Table \ref{tab:nRMSE}, the  $\mathtt{nRMSE}$  values, demonstrate the strong performance of the $\mathtt{ELM}$ model across different energy sources. The consistently low  $\mathtt{nRMSE}$  values underscore the model precision in forecasting. Notably, the error in total consumption “Total” column) shows a very low  $\mathtt{nRMSE}$ ($<0.0511$ for $h=10$), indicating the high potential of $\mathtt{ELM}$ for efficient electrical demand management.
\begin{small}
\begin{longtable}[t]{@{}cccccccc@{}} 
    \caption{$\mathtt{nRMSE}$ for Energy Sources (Unitless) computed for 1 year of predictions (out-sample)} \\ 
    \toprule
    Horizon ($h$) & Total & Thermal & Hydraulics & Solar & Wind & Bioenergy & Imported \\ 
    \midrule
    \endfirsthead
    \bottomrule
    \endfoot
    1 & 0.0219 & 0.0510 & 0.2230 & 0.1791 & 0.4216 & 0.1218 & 0.1368 \\
    2 & 0.0324 & 0.0787 & 0.3292 & 0.3026 & 0.6313 & 0.1981 & 0.2037 \\
    3 & 0.0389 & 0.0952 & 0.3797 & 0.3734 & 0.7658 & 0.2541 & 0.2380 \\
    4 & 0.0430 & 0.1062 & 0.4072 & 0.4178 & 0.8635 & 0.2994 & 0.2581 \\
    5 & 0.0456 & 0.1146 & 0.4233 & 0.4436 & 0.9393 & 0.3377 & 0.2693 \\
    6 & 0.0476 & 0.1208 & 0.4344 & 0.4589 & 0.9981 & 0.3705 & 0.2760 \\
    7 & 0.0488 & 0.1254 & 0.4425 & 0.4666 & 1.0513 & 0.3991 & 0.2801 \\
    8 & 0.0498 & 0.1289 & 0.4488 & 0.4700 & 1.0947 & 0.4244 & 0.2828 \\
    9 & 0.0505 & 0.1317 & 0.4536 & 0.4716 & 1.1322 & 0.4468 & 0.2845 \\
    10 & 0.0511 & 0.1342 & 0.4578 & 0.4722 & 1.1623 & 0.4667 & 0.2861 
    \label{tab:nRMSE}
\end{longtable}
\end{small}
\subsection{Model Comparison with Persistence}
\begin{figure}
\centering
\includegraphics[width=0.95\textwidth]{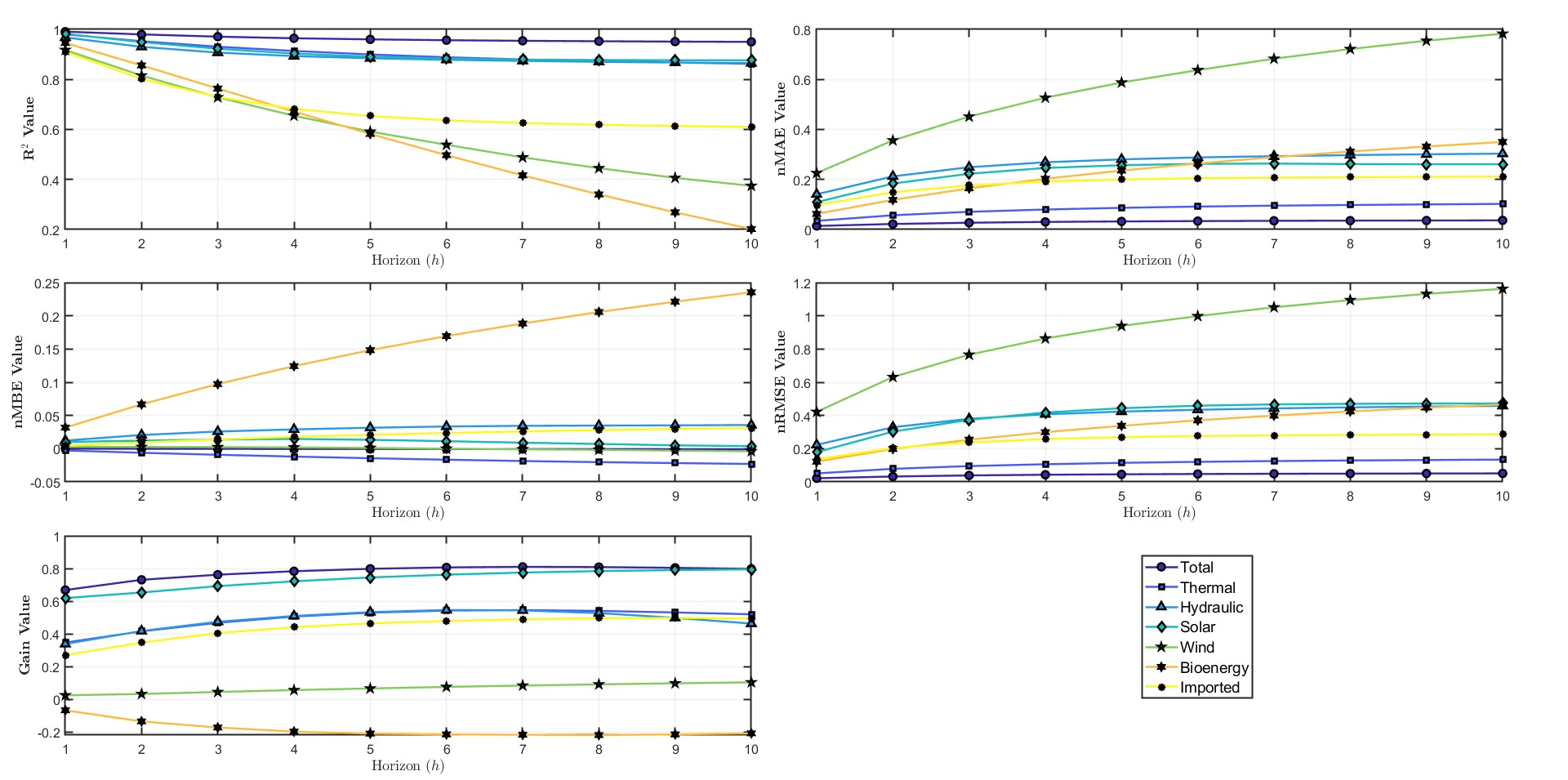}
\caption{Error Metrics Regarding the Horizon of Prediction.}
\label{fig:comp}
\end{figure}

Table \ref{tab:Gain} summarizes the $\mathtt{G}$ metrics computed over 1 year of predictions using out-of-sample data (distinct from the training set), highlighting the significant performance improvements achieved by $\mathtt{ELM}$ over traditional persistence methods. 
These consistently positive Gain  ($\mathtt{G}>0$) for each energy source indicate that $\mathtt{ELM}$ significantly improves forecast accuracy. This improvement is particularly valuable because it supports more informed decision-making, improves operational efficiency, and enhances the reliability of energy supply.
\begin{small}
\begin{longtable}[t]{@{}cccccccc@{}} 
    \caption{$\mathtt{G}$ for Energy Sources (Unitless)} \\ 
    \toprule
    Horizon ($h$) & Total & Thermal & Hydraulic & Solar & Wind & Bioenergy & Imported \\ 
    \midrule
    \endfirsthead
    \bottomrule
    \endfoot
    1 & 0.6681 & 0.3463 & 0.3382 & 0.6188 & 0.0239 & -0.0678 & 0.2691 \\
    2 & 0.7306 & 0.4154 & 0.4174 & 0.6530 & 0.0320 & -0.1360 & 0.3462 \\
    3 & 0.7619 & 0.4677 & 0.4738 & 0.6920 & 0.0441 & -0.1733 & 0.4042 \\
    4 & 0.7831 & 0.5056 & 0.5094 & 0.7216 & 0.0559 & -0.1985 & 0.4407 \\
    5 & 0.7980 & 0.5294 & 0.5329 & 0.7450 & 0.0657 & -0.2106 & 0.4641 \\
    6 & 0.8065 & 0.5424 & 0.5459 & 0.7621 & 0.0754 & -0.2154 & 0.4790 \\
    7 & 0.8099 & 0.5454 & 0.5435 & 0.7750 & 0.0834 & -0.2178 & 0.4892 \\
    8 & 0.8086 & 0.5404 & 0.5269 & 0.7841 & 0.0907 & -0.2196 & 0.4957 \\
    9 & 0.8038 & 0.5308 & 0.4979 & 0.7903 & 0.0973 & -0.2154 & 0.4975 \\
    10 & 0.7975 & 0.5197 & 0.4626 & 0.7945 & 0.1033 & -0.2089 & 0.4948 
    \label{tab:Gain}
\end{longtable}
\end{small}
As such, $\mathtt{ELM}$ seems to represent a significant advance in energy forecasting, presenting a distinct advantage over conventional approaches.
In Figure \ref{fig:comp}, the predicted results for each of the six energy sources are presented, as well as the total electricity consumption of Corsica, over a forecast horizon ranging from 1 to 10 hours. $\mathtt{G}$ provide valuable information on the performance of $\mathtt{ELM}$ over different energy sources and forecast intervals. From this analysis, we can draw several key observations:
The gain in total energy forecasting increases from 0.6681 at a 1-hour horizon, peaking at 0.8099 at 7 hours before slightly declining, suggesting that 7 hours is the optimal forecasting \cyr{horizon}. A similar trend is observed for thermal energy, where the gain rises from 0.3463 to 0.5454 at 7 hours, and for hydraulic energy, which improves from 0.3382 to 0.5459 at 6 hours before a minor decrease.
For solar energy, the gain consistently increases from 0.6188 to 0.7945 over the entire 10-hour horizon, highlighting strong predictive accuracy or alternatively, emphasizing the poor performance of a persistence-based approach. Wind energy, though lower in gains, shows steady improvement from 0.0239 at 1 hour to 0.1033 at 10 hours.
In contrast, bioenergy presents negative gains, starting at -0.0678 and decreasing further to -0.2196 by 8 hours, indicating persistent forecasting challenges. Finally, imported energy sees a steady gain increase from 0.2691 at 1 hour to 0.4975 at 9 hours, where it stabilizes, suggesting sustained improvement in forecast precision.
In summary, the $\mathtt{ELM}$ provides significant improvements with respect to persistence for most energy sources, especially for total consumption, solar, thermal and hydro. However, bioenergy forecasting remains challenging and requires further refinement, even with low installed capacity close to 20 MW for. 

\subsection{Comparison between \texttt{ELM} and \texttt{LSTM}}
\label{res:LSTM}
One may question whether the Extreme Learning Machine ($\mathtt{ELM}$) is better suited for energy demand and supply forecasting than usual deep learning methods, particularly in $\mathtt{MIMO}$ context. According to the literature, these approaches often outperform simpler models.
However, we argue that this trend is not always sustainable, as increasing model complexity can reduce interpretability without necessarily improving performance meaningfully.
In this section, we compare the simple, linear-like $\mathtt{ELM}$ model to the deep Long Short-Term Memory ($\mathtt{LSTM}$) approach. In \ref{annexB}, we detail the theoretical complexity differences between these methods, but it is also crucial to evaluate them in a real-world setting. To this end, we perform experiments on an Intel Core i7-1365U (13th generation) 1.80 GHz processor with 16 GB of RAM.
For the $\mathtt{ELM}$ model with 1,413,120 parameters, training is completed in 104.1 seconds per initialization. In contrast, the $\mathtt{LSTM}$ model (with \texttt{tanh} activation for cell state, \texttt{sigmoid} activation for gates, Adam optimizer, and Mean Squared Error ($\mathtt{MSE}$) loss, trained with \texttt{MaxEpochs} = 5, mini-batch gradient descent, \texttt{MiniBatchSize} = 256, and no explicit early stopping) has 500 hidden neurons and 1,023,507 parameters, requiring 2651.6 seconds for training (approx. 25 times longer). On the same machine, training an $\mathtt{LSTM}$ model with 4096 hidden neurons would be infeasible within a reasonable time frame.
Although deep learning approaches are often powerful, they are also difficult to interpret (users rarely understand the inner workings of the learning process and mechanisms involved) and demand significant computational resources. $\mathtt{ELM}$, being a linear model, offers a straightforward, closed-form solution (without iterative optimization), while $\mathtt{LSTM}$ results may vary depending on the software used (e.g., $\mathtt{Python}^{\text{\textregistered}}$ vs. $\mathtt{Matlab}^{\text{\textregistered}}$  implementations). Moreover, $\mathtt{ELM}$ can be deployed in an online learning context, whereas $\mathtt{LSTM}$ can be unsuitable for real-time applications.
For clarity, we present results for a single forecasting horizon ($h+5$), though similar conclusions apply to other horizons. Table \ref{tab:results} summarizes the performance metrics.
It is evident that these results are directly linked to the limited number of hidden neurons in our $\mathtt{LSTM}$ implementation. This highlights the importance of considering computational constraints when choosing forecasting methods. While deep learning approaches are undeniably powerful, they may not always be practical. In our specific setup, the results obtained with $\mathtt{ELM}$ are consistently more promising than those from $\mathtt{LSTM}$  simulations.
\begin{small}
\begin{table}
    \centering
    \caption{Performance of $\mathtt{LSTM}$ for 5-hour horizon to Compare to Tables \ref{tab:R2} to \ref{tab:Gain}}
    \label{tab:results}
    \begin{tabular}{lcccc}
        \toprule
        \textbf{Category} & $\mathtt{nRMSE}$ & $\mathtt{G}$ & $\mathtt{nMAE}$ & $ \mathtt{R^2} $ \\
        \midrule
        Total   & 0.1722  & 0.2379  & 0.1295  & 0.4336 \\
        Thermal   & 0.3373  & -0.3854  & 0.2593  & 0.1323 \\
        Hydraulic   & 1.1244  & -0.2407  & 0.9957  & 0.1868 \\
        Solar   & 1.3083  & 0.2478  & 1.0437  & 0.0516 \\
        Wind   & 1.5064  & -0.4984  & 1.3293  & -0.0506 \\
        Bioenergy   & 1.5296  & -4.4837  & 1.4164  & -7.5716 \\
        Imported   & 0.4664  & 0.0720  & 0.3611  & -0.0379 \\
        \bottomrule
    \end{tabular}
\end{table}
\end{small}

\subsection{Comparison between \texttt{MIMO} versus \texttt{SISO}}
The objective of this section is to quantify the trade-off between computational speed and predictive performance when comparing the $\mathtt{MIMO}$ and $\mathtt{SISO}$ approaches. The $\mathtt{MIMO}$ model leverages cross-variable interactions, potentially improving accuracy, whereas the $\mathtt{SISO}$ approach treats each variable independently.
If the $\mathtt{MIMO}$ model operates in 104.1 seconds (see Section \ref{res:LSTM}), the $\mathtt{SISO}$ approach requires a longer execution time (240.7 seconds for the 7 energy data). This increase is primarily due to the independent training of multiple models for each variable. The results, presented in Fig. \ref{fig:siso}, illustrate $\mathtt{SISO}$ performance for prediction horizons ranging from 1 hour, passing through 5 hours, up to 10 hours.
Both approaches utilize the same $\mathtt{ELM}$ architecture and analytical resolution method. However, in the $\mathtt{SISO}$ case, a strictly univariate framework is considered (see \ref{annexB}), treating each variable separately without leveraging cross-variable dependencies. This distinction highlights the potential efficiency and accuracy benefits of the $\mathtt{MIMO}$ approach.
\begin{table}
    \centering
    \caption{Comparison of Forecasting Errors ($\mathtt{nRMSE}$ and $\mathtt{nMAE}$) for $\mathtt{SISO}$ Approach at Different Horizons (1h, 5h, and 10h) with Relative Difference to $\mathtt{MIMO}$ (in Parentheses). A Positive Value Indicates Higher Error in $\mathtt{SISO}$ Compared to $\mathtt{MIMO}$, Meaning $\mathtt{MIMO}$ is More Accurate.}
    \label{fig:siso}
    \footnotesize
    \begin{tabular}{lcccccc}
        \toprule
        \multirow{2}{*}{\textbf{Variable}} & \multicolumn{3}{c}{\textbf{nRMSE}} & \multicolumn{3}{c}{\textbf{nMAE}} \\
        \cmidrule(lr){2-4} \cmidrule(lr){5-7}
        & \textbf{1h} & \textbf{5h} & \textbf{10h} & \textbf{1h} & \textbf{5h} & \textbf{10h} \\
        \midrule
        Total     & 0.0219 (0.0\%)  & 0.0467 (2.4\%)  & 0.0520 (1.8\%)  & 0.0143 (0.7\%)  & 0.0320 (-0.9\%)  & 0.0363 (-0.5\%)  \\
        Thermal   & 0.0517 (1.4\%)  & 0.1163 (1.5\%)  & 0.1351 (0.7\%)  & 0.0330 (-4.3\%)  & 0.0859 (-0.9\%)  & 0.1003 (-2.1\%)  \\
        Hydraulic & 0.2307 (3.5\%)  & 0.4403 (4.0\%)  & 0.4732 (3.4\%)  & 0.1470 (4.3\%)  & 0.2930 (4.5\%)  & 0.3118 (2.8\%)  \\
        Solar     & 0.1836 (2.5\%)  & 0.4555 (2.7\%)  & 0.4851 (2.7\%)  & 0.1095 (0.3\%)  & 0.2435 (-5.3\%)  & 0.2463 (-5.6\%)  \\
        Wind      & 0.4221 (0.1\%)  & 0.9514 (1.3\%)  & 1.1786 (1.4\%)  & 0.2289 (1.5\%)  & 0.6344 (7.9\%)  & 0.8744 (11.7\%)  \\
        Bioenergy & 0.1172 (-3.8\%)  & 0.3219 (-4.7\%)  & 0.4476 (-4.1\%)  & 0.0543 (-12.8\%)  & 0.2199 (-6.7\%)  & 0.3322 (-5.1\%)  \\
        Imported  & 0.1384 (1.2\%)  & 0.2770 (2.9\%)  & 0.2925 (2.2\%)  & 0.0977 (0.1\%)  & 0.2007 (0.2\%)  & 0.2098 (-0.7\%)  \\
        \bottomrule
    \end{tabular}
\end{table}
As expected, the prediction error ($\mathtt{SISO}$) increases as the forecast horizon extends. The total energy prediction  remains the most stable across all horizons, with only a moderate increase in error. Thermal and hydroelectric energy also show reasonable predictability, although their errors grow significantly after 5 hours. Wind and bioenergy exhibit the highest errors, with, in first case, an $\mathtt{nRMSE}$ of 1.17 at 10 hours, highlighting the challenge of forecasting these highly variable energy sources. 
Overall, the trend suggests that while the model maintains acceptable accuracy up to 5 hours, performance deteriorates beyond this point, particularly for $\mathtt{RE}$ sources. The performance of the $\mathtt{SISO}$ and $\mathtt{MIMO}$ models was evaluated based on the normalized root-mean-square error ($\mathtt{nRMSE}$) across different forecasting horizons. The results for the $\mathtt{MIMO}$ approach are presented in Figure~\ref{tab:nRMSE}. 
Overall, the $\mathtt{MIMO}$ approach provides a slight improvement over $\mathtt{SISO}$, particularly for hydraulics and solar energy, with error reductions of up to 3.9\% and 2.7\% at 5h and 10h, respectively. The total energy prediction remains stable across both models, while wind energy remains the most challenging variable to forecast, with minimal improvement. 
However, the $\mathtt{MIMO}$ model underperforms for bioenergy, showing an increase in $\mathtt{nRMSE}$ of up to 4.9\% at 5h. This suggests that bioenergy exhibits lower cross-variable dependency, making the additional complexity of $\mathtt{MIMO}$ less beneficial for this particular source.
In conclusion, while $\mathtt{MIMO}$ marginally enhances accuracy for most variables, its gains remain moderate. The choice between $\mathtt{SISO}$ and $\mathtt{MIMO}$ should consider the trade-off between computational cost and forecast accuracy, especially for highly volatile sources like wind and bioenergy. 
\section{Conclusion and Practical Implications}
\label{Sec:Conclusion}
This study addresses key challenges in energy forecasting: non-stationarity, the integration of multiple energy sources, and variable consumption patterns. Using $\mathtt{ELM}$, the approach eliminates the need for extensive preprocessing while ensuring a balance between predictive accuracy and computational efficiency. However, challenges remain, particularly for solar energy, where $\mathtt{nRMSE}>40\%$, reflecting the inherent difficulty of forecasting $\mathtt{PV}$ power generation due to the fact that both installed capacity and available power fluctuate over time. Temporal indices and periodic patterns nonetheless allow the model to adapt to seasonal variations.

As expected, forecast errors increase with longer prediction horizons. The results confirm that $\mathtt{ELM}$ ($\mathtt{MIMO}$) provides stable forecasts for total energy consumption, with moderate error growth. Thermal and hydroelectric generation remain predictable, though uncertainty increases beyond five hours, while renewable energy sources, particularly wind, introduce greater volatility. The $\mathtt{MIMO}$ approach outperforms $\mathtt{SISO}$, particularly for hydro and solar power.
Beyond predictive accuracy, $\mathtt{ELM}$ also surpasses deep learning methods such as $\mathtt{LSTM}$. While $\mathtt{LSTM}$ captures complex temporal dependencies, it suffers from high computational costs, limited interpretability, and sensitivity to software implementation. In contrast, $\mathtt{ELM}$ provides a closed-form solution without iterative optimization, making it better suited for real-time applications, including online learning.
Accurate and rapid forecasting supports energy dispatch optimization. 
While this study demonstrates the effectiveness of $\mathtt{ELM}$ in short-term forecasting, some limitations remain. The model does not explicitly account for operational constraints such as turbine maintenance, hydro reserves, or cost-based dispatch priorities. Future improvements should integrate these factors to enhance experimental applicability. Additionally, an improved $\mathtt{MIMO}$ approach that simultaneously predicts all forecast horizons while increasing the number of model outputs will be very interesting. This could better capture temporal dependencies between successive time steps, improving overall forecasting performance, albeit at a higher computational cost. Hybrid methods combining $\mathtt{ELM}$ with deep learning could also be explored to further enhance long-term predictions and handle highly volatile energy sources.
As power systems transition toward decentralized and renewable-driven models, real-time prediction and optimization will become essential. The proposed methodology provides a valuable decision-support tool for grid operators, helping to balance economic, technical, and environmental objectives. Future research should explore its integration into broader energy management frameworks, ensuring its adaptability to evolving grid infrastructures.

\section*{Acknowledgements}
The authors wish to express their gratitude to $\mathtt{EDF}$, the French grid manager, for providing the electrical data for Corsica, which was essential for conducting this study. 
This research was partially funded by the $\mathtt{ANR}$ under grant $\mathtt{Fine4CAST}$ project ($\mathtt{ANR}$ reference: $\mathtt{22-PETA-0008}$), whose support is gratefully acknowledged. 

\section*{CRediT Authorship Contribution Statement}
CV: Writing – original draft, Methodology, Investigation, Formal analysis, Conceptualization. MD: Writing – original draft, Investigation, Validation, Software. LGG: Investigation, Data curation. MA: Validation, Software, Data curation. YMSD: review \& editing, Methodology. JLD: review \& editing, Methodology. GAF: review \& editing, Methodology EM: review \& editing, Methodology
\appendix

\section{Workflow and Method}
\label{annexA}
\begin{figure}
    \centering
        \begin{tikzpicture}[scale=0.7, transform shape,
        >=latex,
        node distance=1.5cm, 
        every node/.style={draw, rounded corners, align=center, minimum width=2.8cm, minimum height=1cm, font=\footnotesize}, 
        every edge/.style={draw, ->, thick},    ]
        \tikzstyle{input} = [rectangle, fill=gray!20, line width=1pt]
        \tikzstyle{hidden} = [circle, fill=gray!40, line width=1pt]
        \tikzstyle{output} = [rectangle, fill=gray!60, line width=1pt]
        \node[input] (input) {Input Data ($\mathtt{Tot, T, H, S, W, B, I,}$ Time Indexes)};
        \node[input, below=of input] (windowing) {Data Windowing (48 steps)};
        \node[hidden, below=of windowing] (hidden) {Hidden Layer \\ (N = 4096)};
        \node[output, below=of hidden] (output) {Output Layer ($\mathtt{\hat{Tot}, \hat{T}, \hat{H}, \hat{S}, \hat{W}, \hat{B}, \hat{I }}$ \textit{i.e.} Predictions for $h$-horizon)};
        \draw[->] (input) -- (windowing) node[midway, right] {Segmentation, Normalization \\ and Feature Scaling};
        \draw[->] (windowing) -- (hidden) node[midway, right] {ReLU \\ ($\mathcal{H} = \max(0, \mathrm{X}\, \mathcal{W} + b)$)};
        \draw[->] (hidden) -- (output) node[midway, right] {Linear Output \\ ($\mathrm{\hat{Y}} = \mathcal{H} \, \mathcal{W}_{out}$)};
        \node[right=1.72cm of input] {\textbf{\textit{Raw Data}}};
        \node[right=3.1cm of windowing] {\textbf{\textit{Reshaping \& Data Preprocessing}}};
        \node[right=3.9cm of hidden] {\textbf{\textit{Feature Extraction}}};
        \node[right=0.6cm of output] {\textbf{\textit{Output Generation}}};
        \end{tikzpicture}
    \caption{Architecture of an Extreme Learning Machine ($\mathtt{ELM}$) Used for Time Series Forecasting in the Context of Multi-Source Energy Regression Tasks.}
    \label{fig:method_architecture}
\end{figure}
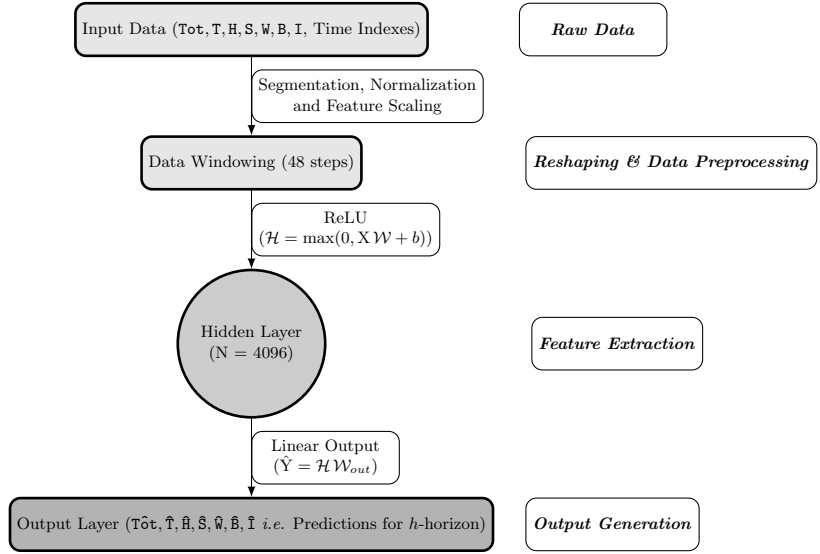

The detail of the predictive method is given in Figure \ref{fig:method_architecture}. The input data comprises total energy consumption ($\mathtt{Tot}$), as well as subcategories including thermal production ($\mathtt{T}$), hydraulics ($\mathtt{H}$), solar ($\mathtt{S}$), wind ($\mathtt{W}$), biomass ($\mathtt{B}$), and import ($\mathtt{I}$), alongside transformed temporal indices in sine/cosine form to capture cyclical variations (hours).
\cyr{The preprocessing step divides the time series into overlapping chunks using a sliding window of 48 steps, where each window advances by a single time step, thus generating a multivariate data point at each shift.} The model applies a $\mathtt{ReLU}$ (rectified linear unit) activation, defined by $\mathcal{H} = \max(0, \mathrm{X}\, \mathcal{W} + b)$, where $\mathrm{X}$ is the segmented input, $\mathcal{W}$ is the weight matrix linking input to the hidden layer, and $b$ is the bias term for each neuron.
The hidden layer (composed of 4096 neurons) and the output are related by linear regression on the activated hidden state ($\mathcal{H}$), described by $\mathrm{\hat{Y}} = \mathcal{H} \, \mathcal{W}_{out}$, where $W_{out}$ is the optimized output weight matrix derived by least squares. 
The output layer predicts future values of energy consumption and production $\mathtt{\hat{Tot}}, \mathtt{\hat{T}}, \mathtt{\hat{H}}, \mathtt{\hat{S}}, \mathtt{\hat{W}}, \mathtt{\hat{B}}, \mathtt{\hat{I}}$ at a forecast horizon of $h$.
This architecture optimizes the model by minimizing the root-mean-square error ($\mathtt{RMSE}$) between predictions and observed values. The windowed data approach allows for robust temporal modeling while maintaining granularity over different types of energy production. The model is tested with $50$ different initializations to optimize the quality of the forecasts thanks to the regularization of the weights.

\section{Extreme Learning Machine (\texttt{ELM}) Architecture in \texttt{MIMO} Case}
\label{annexB}
\begin{figure}
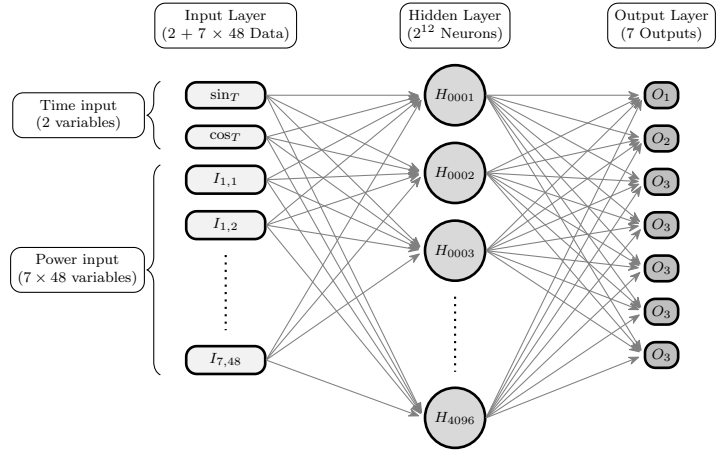

    \centering
    \include{elm}
    \caption{Architecture of an Extreme Learning Machine ($\mathtt{ELM}$) for Time Series Forecasting.}
    \label{fig:ELM_architecture}
\end{figure}

The architecture of the $\mathtt{ELM}$ model, illustrated in Fig. \ref{fig:ELM_architecture}, follows a straightforward design. The input layer consists of seven variables, each carrying 48 past values, plus two time-based features, totaling 338 inputs. These are passed to a hidden layer with $2^{12}$ neurons (4096), which extracts meaningful patterns, before being mapped to seven predicted values at the output layer.
The number of parameters is structured as follows: the input to hidden layer weights total $338 \times 4096 = 1,384,448$, while the hidden to output layer weights add $4096 \times 7 = 28,672$, leading to a total of 1,413,120 parameters.
Despite this high parameter count, the $\mathtt{ELM}$ model remains computationally efficient. Unlike traditional deep learning models, it randomly initializes input weights and does not update them, significantly reducing training complexity. The only computational burden lies in computing the pseudo-inverse of the hidden layer output, leading to a training complexity of $\mathcal{O}(H \times n)$, where $H = 4096$ is the number of hidden neurons and $n$ is the number of training samples. The prediction complexity is given by $\mathcal{O}(H \times (I + O))$, where $I = 338$ inputs and $O = 7$ outputs, yielding $\mathcal{O}(1,412,800)$.
In contrast, Long Short-Term Memory ($\mathtt{LSTM}$) networks, widely used for time-series forecasting \cite{10.1162/neco.1997.9.8.1735}, require iterative weight updates, significantly increasing computational cost. Their training complexity is $\mathcal{O}(n \times H^2)$ due to recurrent loops and gating mechanisms, leading to longer training times. Prediction complexity follows $\mathcal{O}(H^2)$, meaning it also scales quadratically with the number of hidden units.
To quantify the efficiency difference, the ratio of training complexities is given by $\mathcal{O}_{\text{ELM, train}} / \mathcal{O}_{\text{LSTM, train}} = 1 / H$, showing that $\mathtt{ELM}$ requires orders of magnitude fewer computations than $\mathtt{LSTM}$ ($\approx 10^{-4}$ in this study). Similarly, the prediction complexity ratio is $\mathcal{O}_{\text{ELM, predict}} / \mathcal{O}_{\text{LSTM, predict}} = (I + O) / H$, indicating that $\mathtt{ELM}$ is faster at prediction than $\mathtt{LSTM}$, although the advantage is less pronounced compared to training ($\approx 10^{-1}$ in this case).
While $\mathtt{LSTM}$ models are capable of capturing complex temporal dependencies, their computational cost remains significant due to their recurrent structure and gating mechanisms \cite{Goodfellow-et-al-2016}. In contrast, $\mathtt{ELM}$ offers a fast, scalable alternative with minimal training effort and efficient predictions, making it an excellent choice for real-time applications such as energy forecasting and grid management.

\section{Some Prediction Examples}
\label{annexC}
\begin{figure}
\centering
\includegraphics[width=0.7\textwidth]{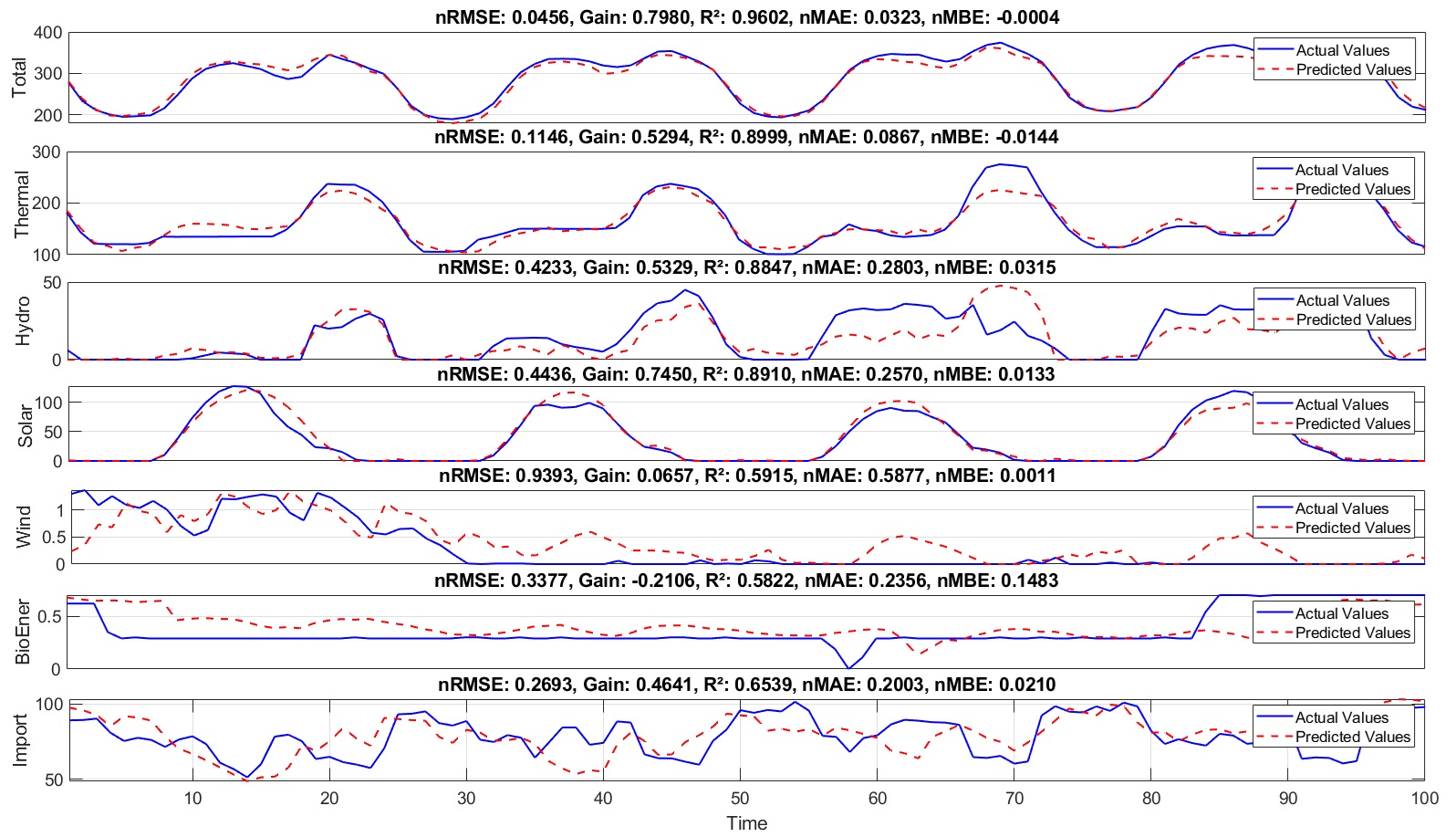}
\caption{Profile of the 7 Predictions (5h horizon) for $\sim$ 4 Consecutive Days (100h) in Winter. Electrical Variables in MW and time in hour.}
\label{fig:h+5}
\end{figure}

\begin{figure}
\centering
\includegraphics[width=0.7\textwidth]{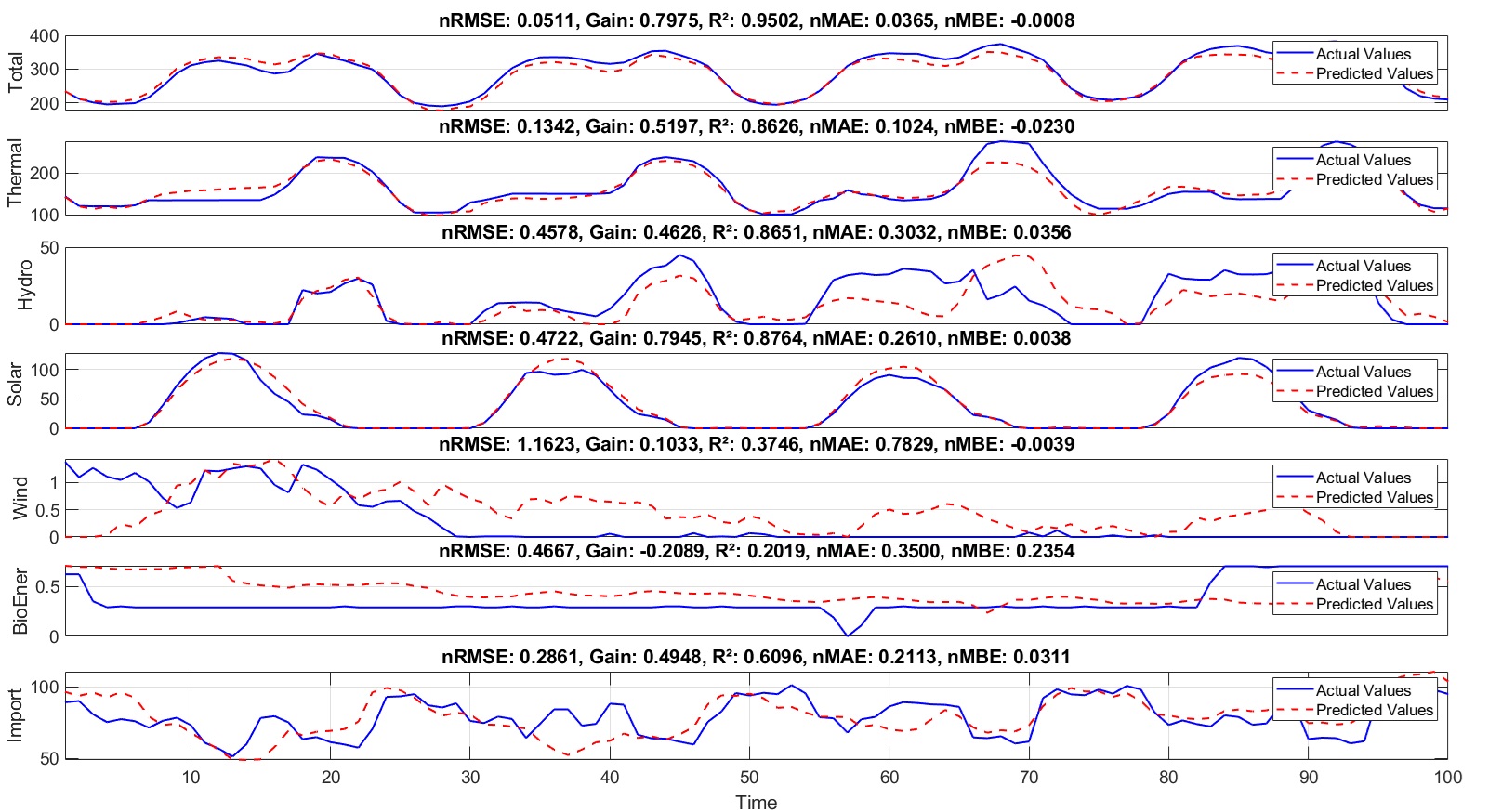}
\caption{Profile of the 7 Predictions (10h horizon) for $\sim$ 4 Consecutive Days (100h) in Winter. Electrical Variables in MW and time in hour.}
\label{fig:h+10}
\end{figure}

\begin{figure}
\centering
\includegraphics[width=0.7\textwidth]{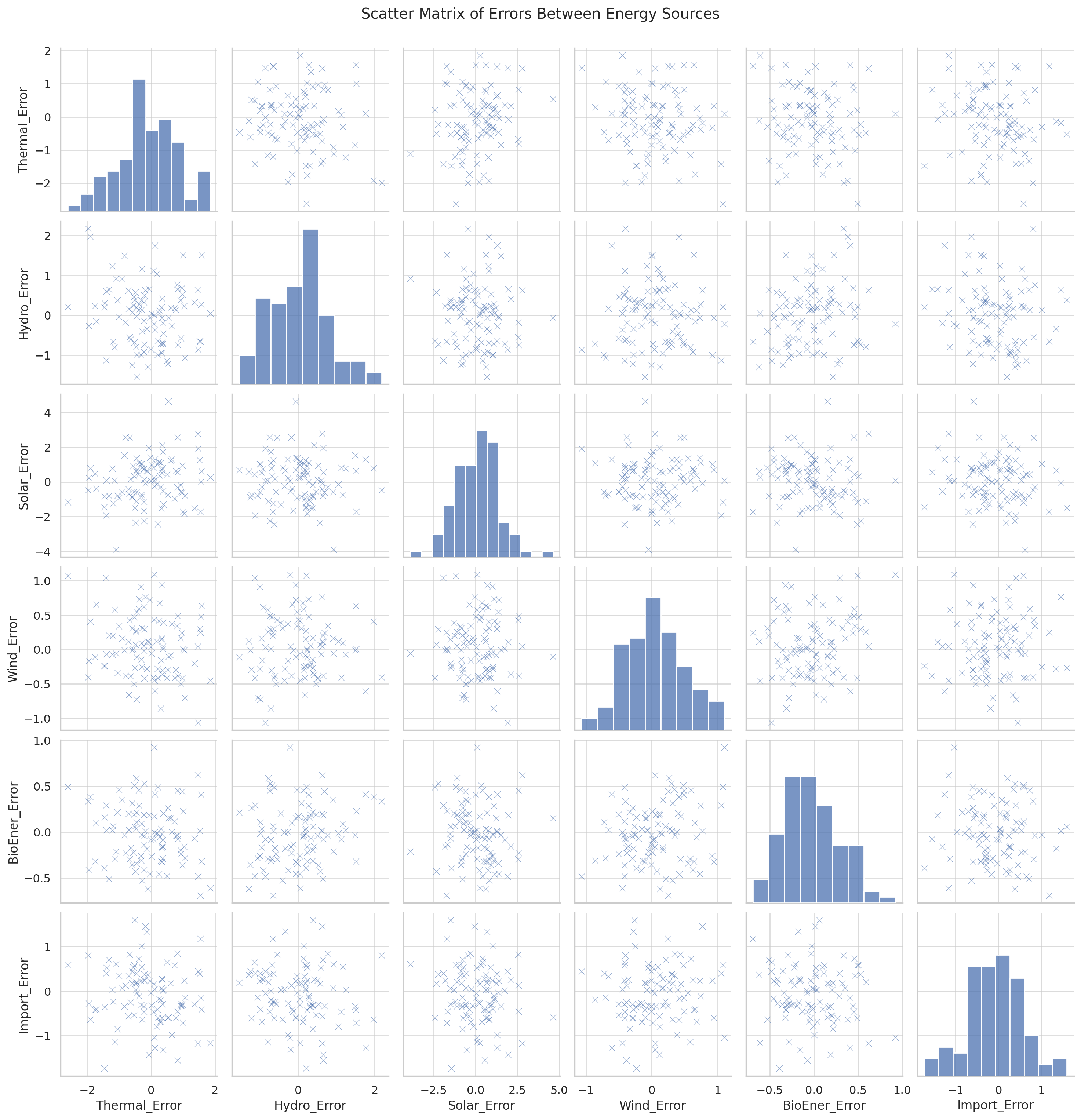}
\caption{Correlation Analysis of Forecast Errors Across Energy Production Sources Over the 4 Days Presented in Figures \ref{fig:h+5} and \ref{fig:h+10}, with Results and Trends Generalizable to the Entire Dataset.}
\label{output(3)}
\end{figure}

This section presents results for the 5h and 10h horizons (Figures \ref{fig:h+5} and \ref{fig:h+10}). In both cases, there’s an issue around the 65-70th hour. When thermal output is underestimated, hydro is overestimated, yet total consumption remains accurate. Forecasting across multiple sources proves challenging; the 5-10h horizon might reveal the limits of this method and, more broadly, the limits of using time-series methods for such forecasts. From these figures, there is an impression that some anti-correlations are present between errors from thermal and hydro, for example, around the 70th hour. To verify this assumption, a scatter plot of the errors for each part of the energy mix is presented in Figure \ref{output(3)}. As visible, there is, in fact, no clear feature, making the validation of statistical dependencies between errors challenging.

\bibliographystyle{elsarticle-num}
\bibliography{Main}
\end{document}

%% file: elm.tex
\begin{tikzpicture}[scale=0.7, transform shape,
    >=latex,
    node distance=0.3cm and 3cm, 
    every node/.style={ align=center, font=\scriptsize}]

    \tikzstyle{input} = [draw, rounded corners,rectangle, fill=gray!10, line width=1pt, minimum width=1.5cm]
    \tikzstyle{hidden} = [draw, circle, fill=gray!30, line width=1pt]
    \tikzstyle{output} = [draw, rounded corners, rectangle, fill=gray!50, line width=1pt]
    \tikzstyle{arrow} = [->, gray, shorten >=2pt, >={Stealth[round]}]
    \tikzstyle{dots} = [dotted, shorten <=0.2cm, shorten >=0.2cm, thick]
    \tikzstyle{brace} = [decorate, decoration={brace, mirror, amplitude=5pt, raise=2ex}]
    \tikzstyle{header} = [draw, rounded corners,rectangle]
    
    \node[input] (i1) {$\text{sin}_T$};
    \node[input, below= of i1] (i2) {$\text{cos}_T$};
    \node[input, below= of i2] (i3) {$I_{1,1}$};
    \node[input, below= of i3] (i4) {$I_{1,2}$};
    \node[input, below= 2 cm of i4] (iN) {$I_{7,48}$};
    \draw[dots] (i4) -- (iN);

    \node[hidden, right=of i1] (h1) {$H_{0001}$};
    \node[hidden, below=of h1] (h2) {$H_{0002}$};
    \node[hidden, below=of h2] (h3) {$H_{0003}$};
    \node[hidden, below=2cm of h3] (hN) {$H_{4096}$}; 
    \draw[dots] (h3) -- (hN);

    \node[output, right=of h1] (o1) {$O_1$};      
    \node[output, below=of o1] (o2) {$O_2$};  
    \node[output, below=of o2] (o3) {$O_3$};  
    \node[output, below=of o3] (o4) {$O_3$};  
    \node[output, below=of o4] (o5) {$O_3$};  
    \node[output, below=of o5] (o6) {$O_3$};  
    \node[output, below=of o6] (o7) {$O_3$};  

    \draw [brace] (i1.north west) -- (i2.south west) 
        node[header, midway, left, xshift=-2em, minimum width=2.5cm]
        {Time input\\(2 variables)};
    \draw [brace] (i3.north west) -- (iN.south west) 
        node[header, midway, left, xshift=-2em, minimum width=2.5cm]
        {Power input\\($7 \times 48$ variables)};
    \node[header, above=of h1](hidden) {Hidden Layer \\ ($2^{12}$ Neurons)};
    \node[header] at (i1|-hidden) {Input Layer \\ (2 + 7 $\times$ 48 Data)};
    \node[header] at (o1|-hidden) {Output Layer \\ (7 Outputs)};

    \foreach \i in {1, 2, ..., 4, N} {
        \foreach \j in {1, 2, ..., 3, N} {
            \draw[arrow] (i\i.east) -- (h\j.west);
        } 
    };

    \foreach \i in {1, 2, ..., 3, N} {
        \foreach \j in {1, 2, ..., 7} {
            \draw[arrow] (h\i.east) -- (o\j.west);
        } 
    };
\end{tikzpicture}

%% file: Main.bbl
\begin{thebibliography}{10}
\expandafter\ifx\csname url\endcsname\relax
  \def\url#1{\texttt{#1}}\fi
\expandafter\ifx\csname urlprefix\endcsname\relax\def\urlprefix{URL }\fi
\expandafter\ifx\csname href\endcsname\relax
  \def\href#1#2{#2} \def\path#1{#1}\fi

\bibitem{moretti2020grid}
A.~Moretti, C.~Pitas, G.~Christofi, E.~Bu{\'e}, M.~G. Francescato, Grid
  integration as a strategy of {med-TSO} in the mediterranean area in the
  framework of climate change and energy transition, Energies 13~(20) (2020)
  5307.
\newblock \href {https://doi.org/10.3390/en13205307}
  {\path{doi:10.3390/en13205307}}.

\bibitem{basit2020limitations}
M.~A. Basit, S.~Dilshad, R.~Badar, S.~M. Sami~ur Rehman, Limitations,
  challenges, and solution approaches in grid-connected renewable energy
  systems, International Journal of Energy Research 44~(6) (2020) 4132--4162.
\newblock \href {https://doi.org/10.1002/er.5033} {\path{doi:10.1002/er.5033}}.

\bibitem{POMMERET2022104273}
A.~Pommeret, K.~Schubert, Optimal energy transition with variable and
  intermittent renewable electricity generation, Journal of Economic Dynamics
  and Control 134 (2022) 104273.
\newblock \href {https://doi.org/10.1016/j.jedc.2021.104273}
  {\path{doi:10.1016/j.jedc.2021.104273}}.

\bibitem{notton2018intermittent}
G.~Notton, M.-L. Nivet, C.~Voyant, C.~Paoli, C.~Darras, F.~Motte, A.~Fouilloy,
  Intermittent and stochastic character of renewable energy sources:
  Consequences, cost of intermittence and benefit of forecasting, Renewable and
  sustainable energy reviews 87 (2018) 96--105.
\newblock \href {https://doi.org/10.1016/j.rser.2018.02.007}
  {\path{doi:10.1016/j.rser.2018.02.007}}.

\bibitem{su15097087}
N.~E. Benti, M.~D. Chaka, A.~G. Semie, Forecasting renewable energy generation
  with machine learning and deep learning: Current advances and future
  prospects, Sustainability 15~(9) (2023).
\newblock \href {https://doi.org/10.3390/su15097087}
  {\path{doi:10.3390/su15097087}}.

\bibitem{TALARI20181953}
S.~Talari, M.~Shafie-khah, G.~J. Osório, J.~Aghaei, J.~P. Catalão, Stochastic
  modelling of renewable energy sources from operators' point-of-view: A
  survey, Renewable and Sustainable Energy Reviews 81 (2018) 1953--1965.
\newblock \href {https://doi.org/10.1016/j.rser.2017.06.006}
  {\path{doi:10.1016/j.rser.2017.06.006}}.

\bibitem{ALAMO2019111}
D.~H. Alamo, R.~N. Medina, S.~D. Ruano, S.~S. García, K.~P. Moustris, K.~K.
  Kavadias, D.~Zafirakis, G.~Tzanes, E.~Zafeiraki, G.~Spyropoulos, J.~K.
  Kaldellis, G.~Notton, J.-L. Duchaud, M.-L. Nivet, A.~Fouilloy, S.~Lespinats,
  An advanced forecasting system for the optimum energy management of island
  microgrids, Energy Procedia 159 (2019) 111--116, {Renewable Energy
  Integration with Mini/Microgrid}.
\newblock \href {https://doi.org/10.1016/j.egypro.2018.12.027}
  {\path{doi:10.1016/j.egypro.2018.12.027}}.

\bibitem{voyant2017machine}
C.~Voyant, G.~Notton, S.~Kalogirou, M.-L. Nivet, C.~Paoli, F.~Motte,
  A.~Fouilloy, Machine learning methods for solar radiation forecasting: A
  review, Renewable energy 105 (2017) 569--582.
\newblock \href {https://doi.org/10.1016/j.renene.2016.12.095}
  {\path{doi:10.1016/j.renene.2016.12.095}}.

\bibitem{teo2015forecasting}
T.~T. Teo, T.~Logenthiran, W.~L. Woo, Forecasting of photovoltaic power using
  extreme learning machine, in: 2015 IEEE Innovative Smart Grid
  Technologies-Asia (ISGT ASIA), IEEE, 2015, pp. 1--6.
\newblock \href {https://doi.org/10.1109/ISGT-Asia.2015.7387113}
  {\path{doi:10.1109/ISGT-Asia.2015.7387113}}.

\bibitem{al2018extreme}
S.~Al-Dahidi, O.~Ayadi, J.~Adeeb, M.~Alrbai, B.~R. Qawasmeh, Extreme learning
  machines for solar photovoltaic power predictions, Energies 11~(10) (2018)
  2725.
\newblock \href {https://doi.org/10.3390/en11102725}
  {\path{doi:10.3390/en11102725}}.

\bibitem{la2021new}
J.~L.~G. La~Salle, M.~David, P.~Lauret, A new climatology reference model to
  benchmark probabilistic solar forecasts, Solar Energy 223 (2021) 398--414.
\newblock \href {https://doi.org/10.1016/j.solener.2021.05.037}
  {\path{doi:10.1016/j.solener.2021.05.037}}.

\bibitem{he2016cyber}
H.~He, J.~Yan, Cyber-physical attacks and defences in the smart grid: a survey,
  IET Cyber-Physical Systems: Theory \& Applications 1~(1) (2016) 13--27.
\newblock \href {https://doi.org/10.1049/iet-cps.2016.0019}
  {\path{doi:10.1049/iet-cps.2016.0019}}.

\bibitem{reda2022comprehensive}
H.~T. Reda, A.~Anwar, A.~Mahmood, Comprehensive survey and taxonomies of false
  data injection attacks in smart grids: attack models, targets, and impacts,
  Renewable and Sustainable Energy Reviews 163 (2022) 112423.
\newblock \href {https://doi.org/10.1016/j.rser.2022.112423}
  {\path{doi:10.1016/j.rser.2022.112423}}.

\bibitem{le2022smart}
T.~D. Le, H.~P.~T. Nguyen, K.-T. Huynh, R.~Beuran, Smart grid cyber-attack
  analysis and countermeasures, in: 2022 RIVF International Conference on
  Computing and Communication Technologies (RIVF), IEEE, 2022, pp. 590--595.
\newblock \href {https://doi.org/10.1109/RIVF55975.2022.10013873}
  {\path{doi:10.1109/RIVF55975.2022.10013873}}.

\bibitem{fuller2020digital}
A.~Fuller, Z.~Fan, C.~Day, C.~Barlow, Digital twin: Enabling technologies,
  challenges and open research, IEEE Access 8 (2020) 108952--108971.
\newblock \href {https://doi.org/10.1109/ACCESS.2020.2998358}
  {\path{doi:10.1109/ACCESS.2020.2998358}}.

\bibitem{bishaw2024}
F.~G. Bishaw, M.~K. Ishak, T.~H. Atyia, Review of artificial intelligence
  applications in renewable energy systems integration, J. Electrical Systems
  20~(3) (2024) 566--582.
\newblock \href {https://doi.org/10.52783/jes.2983}
  {\path{doi:10.52783/jes.2983}}.

\bibitem{benti2023}
N.~E. Benti, M.~D. Chaka, A.~G. Semie, Forecasting renewable energy generation
  with machine learning and deep learning: Current advances and future
  prospects, Sustainability 15~(9) (2023) 7087.
\newblock \href {https://doi.org/10.3390/su15097087}
  {\path{doi:10.3390/su15097087}}.

\bibitem{alansari2023}
M.~S.~A. Ansari, {Utilization of Artificial Intelligence (AI) and Machine
  Learning (ML) in the Field of Energy Research}, International Journal on
  Recent and Innovation Trends in Computing and Communication 11~(3) (2023)
  305--318.
\newblock \href {https://doi.org/10.17762/ijritcc.v11i3.9774}
  {\path{doi:10.17762/ijritcc.v11i3.9774}}.

\bibitem{elsaraiti2024shortterm}
M.~Elsaraiti, Short-term power consumption forecasting using neural networks
  with first- and second-order differencing, Academic Energy Journal 1~(3)
  (2024) 10--20.
\newblock \href {https://doi.org/10.20935/AcadEnergy7381}
  {\path{doi:10.20935/AcadEnergy7381}}.

\bibitem{Saglam2023}
M.~Saglam, C.~Spataru, O.~A. Karaman, Forecasting electricity demand in
  {Turkey} using optimization and machine learning algorithms, Energies 16~(11)
  (2023) 4499.
\newblock \href {https://doi.org/10.3390/en16114499}
  {\path{doi:10.3390/en16114499}}.

\bibitem{Ardabili2022}
S.~Ardabili, L.~Abdolalizadeh, C.~Mako, B.~Torok, Systematic review of deep
  learning and machine learning for building energy prediction, Frontiers in
  Energy Research (2022).
\newblock \href {https://doi.org/10.3389/fenrg.2022.786027}
  {\path{doi:10.3389/fenrg.2022.786027}}.

\bibitem{csse.2023.037735}
A.~Motwakel, M.~Obayya, N.~Nemri, K.~Tarmissi, H.~Mohsen, M.~Rizwanulla,
  I.~Yaseen, A.~S. Zamani, Predictive multimodal deep learning-based
  sustainable renewable and non-renewable energy utilization, Computer Systems
  Science and Engineering 47~(1) (2023) 1267--1281.
\newblock \href {https://doi.org/10.32604/csse.2023.037735}
  {\path{doi:10.32604/csse.2023.037735}}.

\bibitem{Ding2022}
S.~Ding, H.~Zhang, Z.~Tao, R.~Li, Integrating data decomposition and machine
  learning methods: An empirical proposition for renewable energy generation
  forecasting, Expert Systems with Applications 200 (2022) 117025.
\newblock \href {https://doi.org/10.1016/j.eswa.2022.117635}
  {\path{doi:10.1016/j.eswa.2022.117635}}.

\bibitem{Revathi2023}
B.~S. Revathi, A survey on advanced machine learning and deep learning
  techniques assisting in renewable energy generation, Environmental Science
  and Pollution Research (2023).
\newblock \href {https://doi.org/10.1007/s11356-023-29064-w}
  {\path{doi:10.1007/s11356-023-29064-w}}.

\bibitem{ABDELAZIZ20229447}
R.~M. {Abd El-Aziz}, Renewable power source energy consumption by hybrid
  machine learning model, Alexandria Engineering Journal 61~(12) (2022)
  9447--9455.
\newblock \href {https://doi.org/10.1016/j.aej.2022.03.019}
  {\path{doi:10.1016/j.aej.2022.03.019}}.

\bibitem{en11092475}
M.~Li, W.~Wang, G.~De, X.~Ji, Z.~Tan, Forecasting carbon emissions related to
  energy consumption in {Beijing-Tianjin-Hebei} region based on grey prediction
  theory and extreme learning machine optimized by support vector machine
  algorithm, Energies 11~(9) (2018).
\newblock \href {https://doi.org/10.3390/en11092475}
  {\path{doi:10.3390/en11092475}}.

\bibitem{SIBTAIN2022115703}
M.~Sibtain, H.~Bashir, M.~Nawaz, S.~Hameed, M.~{Imran Azam}, X.~Li, T.~Abbas,
  S.~Saleem, A multivariate ultra-short-term wind speed forecasting model by
  employing multistage signal decomposition approaches and a deep learning
  network, Energy Conversion and Management 263 (2022) 115703.
\newblock \href {https://doi.org/10.1016/j.enconman.2022.115703}
  {\path{doi:10.1016/j.enconman.2022.115703}}.

\bibitem{Han2024}
H.~Han, Z.~Liu, M.~Barrios~Barrios, J.~Li, Z.~Zeng, N.~Sarhan, E.~M. Awwad,
  Time series forecasting model for non-stationary series pattern extraction
  using deep learning and garch modeling, Journal of Cloud Computing 13~(1)
  (2024) 2.
\newblock \href {https://doi.org/10.1186/s13677-023-00576-7}
  {\path{doi:10.1186/s13677-023-00576-7}}.

\bibitem{NOTTON20191157}
G.~Notton, J.~Duchaud, M.~Nivet, C.~Voyant, K.~Chalvatzis, A.~Fouilloy, {The
  electrical energy situation of French islands and focus on the Corsican
  situation}, Renewable Energy 135 (2019) 1157--1165.
\newblock \href {https://doi.org/10.1016/j.renene.2018.12.090}
  {\path{doi:10.1016/j.renene.2018.12.090}}.

\bibitem{insee_essentiel_corse_2024}
I.~N. de~la Statistique et des Études~Économiques (INSEE),
  \href{https://www.insee.fr/fr/statistiques/4481069}{L'essentiel sur la
  {Corse}} (octobre 2024).
\newline\urlprefix\url{https://www.insee.fr/fr/statistiques/4481069}

\bibitem{cre_transition_energetique}
{Commission de régulation de l'énergie (CRE)},
  \href{https://www.cre.fr/electricite/transition-energetique-dans-les-zni.html}{Transition
  énergétique dans les zni} (2023).
\newline\urlprefix\url{https://www.cre.fr/electricite/transition-energetique-dans-les-zni.html}

\bibitem{edf_open_data_corse}
E.~Corse, \href{https://opendata-corse.edf.fr/pages/home0/}{Open data {EDF
  Corse}} (2024).
\newline\urlprefix\url{https://opendata-corse.edf.fr/pages/home0/}

\bibitem{GAIRAA2022890}
K.~Gairaa, C.~Voyant, G.~Notton, S.~Benkaciali, M.~Guermoui, Contribution of
  ordinal variables to short-term global solar irradiation forecasting for
  sites with low variabilities, Renewable Energy 183 (2022) 890--902.
\newblock \href {https://doi.org/10.1016/j.renene.2021.11.028}
  {\path{doi:10.1016/j.renene.2021.11.028}}.

\bibitem{10.1007/978-3-642-04138-9_30}
N.~Veyrat-Charvillon, F.-X. Standaert, Mutual information analysis: How, when
  and why?, in: C.~Clavier, K.~Gaj (Eds.), Cryptographic Hardware and Embedded
  Systems - CHES 2009, Springer Berlin Heidelberg, Berlin, Heidelberg, 2009,
  pp. 429--443.
\newblock \href {https://doi.org/10.1007/978-3-642-04138-9_30}
  {\path{doi:10.1007/978-3-642-04138-9_30}}.

\bibitem{10.1162/neco.1997.9.8.1735}
S.~Hochreiter, J.~Schmidhuber, {Long Short-Term Memory}, Neural Computation
  9~(8) (1997) 1735--1780.
\newblock \href {https://doi.org/10.1162/neco.1997.9.8.1735}
  {\path{doi:10.1162/neco.1997.9.8.1735}}.

\bibitem{Goodfellow-et-al-2016}
I.~Goodfellow, Y.~Bengio, A.~Courville, Deep Learning, MIT Press, 2016,
  \url{http://www.deeplearningbook.org}.

\end{thebibliography}
